\documentclass[nonacm,sigconf]{acmart}
\acmConference[The Web Conference]{}{2020}{Taipei}
\usepackage{epsfig}
\usepackage{graphicx}
\usepackage{amsmath}
\usepackage{amssymb}
\usepackage{stackengine}
\usepackage{float}
\usepackage[ruled]{algorithm2e}
\usepackage[noend]{algpseudocode}
\usepackage{tabularx}
\usepackage{soul}
\usepackage{url}
\usepackage{caption}
\newtheorem{thm}{Theorem}[section]
\newcommand{\appendixhead}%
{\centering \huge Appendix
\vspace{0.25in}}

\usepackage{booktabs}
\usepackage{subcaption}
\usepackage{multirow}
\renewcommand\footnotetextcopyrightpermission[1]{} 
\AtBeginDocument{%
  \providecommand\BibTeX{{%
    \normalfont B\kern-0.5em{\scshape i\kern-0.25em b}\kern-0.8em\TeX}}}

\setcopyright{none}

\acmSubmissionID{1647}

\begin{document}

\title{ShapeVis: High-dimensional Data Visualization at Scale}

\author{Nupur Kumari$^*$, Siddarth R.$^*$,  Akash Rupela}
\authornote{authors contributed equally}
\email{nupkumar@adobe.com, sir@adobe.com, arupela@adobe.com}

\author{Piyush Gupta$^*$, Balaji Krishnamurthy}
\email{piygupta@adobe.com, kbalaji@adobe.com}

\affiliation{%
  \institution{Media and Data Science Research lab\\ Adobe Experience Cloud}
}


\begin{abstract}
We present ShapeVis, a scalable visualization technique for point cloud data inspired from topological data analysis. Our method captures the underlying geometric and topological structure of the data in a compressed graphical representation. Much success has been reported by the data visualization technique Mapper, that discreetly approximates the Reeb graph of a filter function on the data. However, when using standard dimensionality reduction algorithms as the filter function, Mapper suffers from considerable computational cost. This makes it difficult to scale to high-dimensional data. Our proposed technique relies on finding a subset of points called landmarks along the data manifold to construct a weighted witness-graph over it. This graph captures the structural characteristics of the point cloud, and its weights are determined using a Finite Markov Chain. We further compress this graph by applying induced maps from standard community detection algorithms. Using techniques borrowed from manifold tearing, we prune and reinstate edges in the induced graph based on their modularity to summarize the shape of data.
We empirically demonstrate how our technique captures the structural characteristics of real and synthetic data sets. Further, we compare our approach with Mapper using various filter functions like t-SNE, UMAP, LargeVis and show that our algorithm scales to millions of data points while preserving the quality of data visualization.
\end{abstract}




\maketitle
\sloppy

\section{Introduction}  
With ever-increasing amounts of data and advances in hardware to store and query such datasets, it has become critical to have scalable and robust systems for analyzing big data. Understanding and mining insights from the massive amount of data has made a significant impact in fields like marketing, business, education and healthcare. However, we continue to lack the tools and techniques to produce insight-generating visualizations of large-scale and high-dimensional data. Traditional visualization approaches like scatter plots and heat-maps have been proven to be effective only for small or intermediate sizes of data. These techniques are extremely intuitive and can be used to determine the bivariate relationships between variables. However, they require laying out the data points on a lower-dimensional space which is computationally intractable when data is large-scale and high-dimensional. Moreover, in some cases, these visualizations can lead to inconclusive results.

Dimensionality reduction is the transformation of high-dimensional data into a low dimensional representation while preserving some desirable structure of data. It is a core problem in machine learning and data mining since, generally, real-world data is found to lie on a low-dimensional manifold embedded in high-dimensional space \cite{manifold_hypothesis}. Of late, the usage of dimensionality reduction for visualization of high-dimensional data has become common practice following the success of techniques such as PCA\cite{pca}, MDS \cite{mds} t-SNE \cite{tsne}, tsNET \cite{tsNET}, and UMAP\cite{umap}. These techniques are being applied in a wide range of fields and on ever-increasing sizes of datasets. Broadly, dimension reduction algorithms tend to fall into two categories. Algorithms such as PCA \cite{pca} and MDS \cite{mds} seek to preserve the distance structure within the data whereas algorithms like t-SNE \cite{tsne}, Isomap \cite{isomap}, LargeVis \cite{largevis}, UMAP \cite{umap} and Laplacian Eigenmaps \cite{leigen} favor the preservation of local distances over global distance.


The class of techniques known as Stochastic Neighbor Embedding (SNE) \cite{SNE} are considered the current state-of-the-art in dimension reduction for visualization. Intuitively, SNE \cite{SNE} techniques encode local relationships in the original and embedding spaces as probability distributions. Then it minimizes the loss of information between the two distributions with respect to the locations of points in the map. SNE-based approaches have revealed many interesting structures in real-world and synthetic data \cite{SNEapps}.

However, there are issues with dimensionality reduction methods. Since these methods compress a large number of attributes down to a few, they suffer from projection losses. As a result, points well separated in high-dimensional space might appear to be in the same neighborhood in the lower-dimensional projection. In particular, due to local neighborhood preservation SNE techniques might miss structures at different sizes. Though SNE is fairly robust to changes in perplexity \cite{tsne}, multiple plots with different perplexities are typically needed to arrive at a conclusively useful embedding \cite{wattenberg2016how}. They are also known to not always produce similar output on successive runs making them hard to use. Moreover, application of SNE techniques to large datasets is problematic, as the computational complexity is usually $\mathcal{O}(n^2)$\cite{tsne}. Using approximations it can be reduced to $\mathcal{O}(n\log(n))$\cite{van2014accelerating}. Even when appropriately projected onto lower-dimensional spaces, interpreting and extracting insights can be cumbersome when there are too many data points or when the data is high-dimensional. This is largely because 2D and 3D representations do not provide any obvious means to explore and understand critical patterns and global relationships in the data.

Visualization approaches that make use of abstractions and concise representations are, therefore, essential in capturing and interpreting structural information in high-dimensional data. In this paper, we propose ShapeVis, that tries to address this issue by computing a compressed representation of large-scale and high-dimensional data in the form of a graph. One important goal of data visualization is to let the user obtain knowledge about the intrinsic structure of data, i.e. to understand how it is organized on a large scale. To this end, our proposed algorithm focuses on preserving global distances and topology. ShapeVis encodes the geometric properties of the original data by capturing small-neighborhood relationships within its nodes and topological features like connected components and loops in the graph structure. ShapeVis first finds a subset of points called landmarks along the data manifold and constructs a weighted graph over it that approximates a $1$-witness complex\cite{deSilva} on the landmarks. A second level of landmarks are selected in such a manner that their neighborhoods partition the witness graph. The weights between these landmarks are determined by modelling the random movement of a hypothetical particle on the landmarks using a Finite Markov Chain similar to hSNE\cite{hsne}. This graph is compressed by applying induced maps from standard community detection algorithms. We then prune and reinstate edges in the induced graph based on their modularity to summarize the shape of data. Given its construction, it is scalable to millions of data-points, and the final visualization graph is easy to analyze.

We perform extensive experiments on real-world, large-scale and high-dimensional data sets, including images, text (word embedding) and networks. Experimental results show that our proposed algorithm for constructing the compressed representation captures concise and intuitive visualizations. We compare our approach to Mapper\cite{mapper} with standard dimensionality reduction functions like LargeVis, UMAP and tSNE as its filter function. We find ShapeVis to be much more efficient when the dataset size is huge. To summarize, we make the following contributions:

\begin{enumerate}
    \item We propose a visualization technique that captures the intrinsic shape of large-scale and high-dimensional data efficiently
    \item We propose a weighting scheme based on a Finite Markov Chain (FMC) built on a witness complex to encode similarities between landmarks
    \item We propose a manifold tearing procedure that captures essential loops and connectivity of the data manifold.
    \item We conduct experiments on large, real-world data
sets and compare ShapeVis with Mapper on LargeVis, UMAP and tSNE, both computationally and visually.
\end{enumerate}

\section{Related Work}\label{relatedwork}

\paragraph{\textbf{Mapper}} Techniques in Topological Data Analysis (TDA) compute and analyze topological features of generally high-dimensional and possibly noisy data sets. In particular, the mapper algorithm \cite{mapper} is the most scalable among TDA approaches. It converts data into a graphical representation where nodes represent collections of similar data points and edges between nodes represent the existence of shared data points between the nodes. It discreetly approximates the Reeb graph\cite{reeb} of a filter function on the manifold. Despite its success in several business problems, Mapper is an ad hoc technique requiring a significant amount of parameter tuning\cite{carriere2018statistical}. Also, Mapper only provides a global scale selection for the covers of data which often results in bad visualizations when data is of non-uniform density. Multimapper \cite{deb2018multimapper} tried to address this issue by a scale selection scheme sensitive to the local density of data points. Moreover, when using standard dimensionality reduction algorithms for filter function, Mapper suffers from considerable computational cost. TDA techniques like Mapper has been applied to a variety of domains namely shape analysis \cite{collins2004barcode}, sensor-network coverage \cite{desilva2007}, proteins \cite{kovacev2016using}, images \cite{carlsson2008local} \cite{letscher2007image}, social network analysis \cite{carstens2013persistent}\cite{patania2017shape}, computational neuroscience \cite{lee2011discriminative}, genomics  \cite{camara2016topological} \cite{camara2017topological}, periodicity in time series \cite{seversky2016time}, cancer \cite{nicolau2011topology}, materials science \cite{buchet2018persistent} \cite{lee2018high}, financial networks \cite{gidea2017topological} and more. 

\paragraph{\textbf{Landmark selection}} Manifold landmarking has been widely used to find a subset of data-points that capture the structural characteristics of the underlying manifold. This helps in reducing the time and space complexity of the subsequent steps in the algorithm. So far, several landmark selection methods have been proposed. Landmarks can be selected randomly from the given data as given in \cite{silva2003global}. Another interesting approach is the landmarking procedure described in Fast-Isomap proposed in \cite{lei2013increasing} based on integer optimization. In \cite{liang2015enhancing}, landmarks are chosen from a central part of the given data.

\paragraph{\textbf{Witness complex}} The witness complex\cite{deSilva} is a computationally feasible approximation of the underlying topological structure of a point cloud. It is built in reference to a subset of points, the landmarks, rather than considering all the points as in the \v{C}ech and Vietoris-Rips complexes \cite{Hatcher}. The witness complex is a weak approximation\cite{guibas2008reconstruction} of the Delaunay triangulation\cite{de2008delaunay} on the data. 

\paragraph{\textbf{Manifold tearing}} Lee and Verleysen \cite{lee2005nonlinear} introduce a manifold tearing procedure to cut out essential loops to aid in downstream dimensionality-reduction. First, a $k$-nearest neighbor (kNN) graph on the point cloud sample is constructed. Then, a minimum spanning tree (MST) or a shortest path tree (SPT) containing no cycles is computed on this kNN graph. Finally, edges not generating non-contractible cycles with more than $4$ edges are reintroduced to form the final torn graph for downstream dimensionality reduction.

\begin{figure*}[t]
\centering
\begin{subfigure}{0.25\linewidth}
    \centering \includegraphics[width=\linewidth]{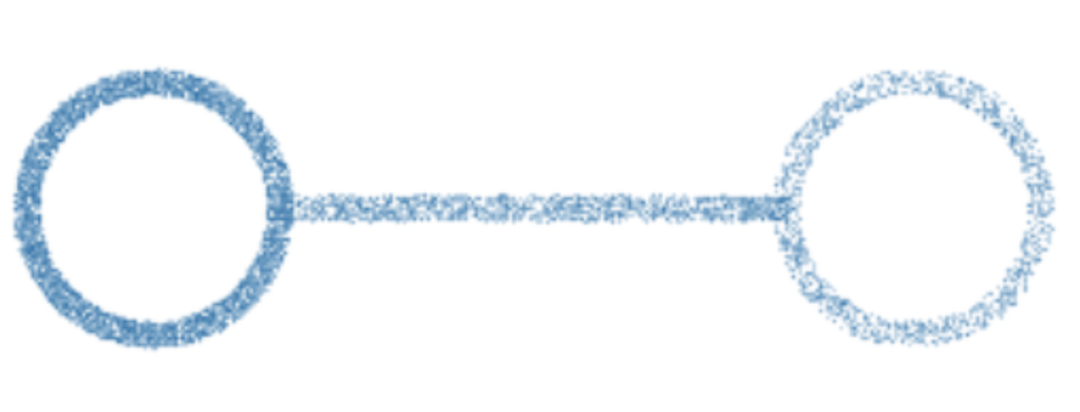} 
     \caption{Point Cloud}
\end{subfigure}%
\begin{subfigure}{0.25\linewidth}
    \centering \includegraphics[width=\linewidth]{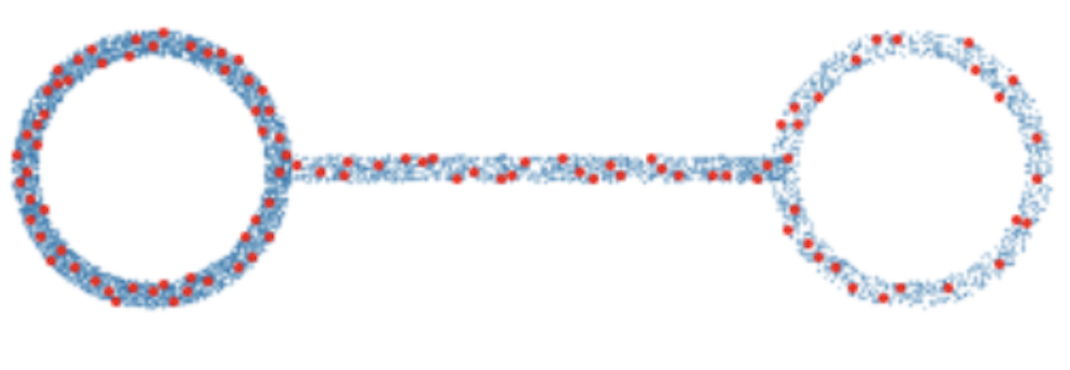} 
     \caption{Landmark Sampling}
\end{subfigure}%
\begin{subfigure}{0.25\linewidth}
    \centering \includegraphics[width=\linewidth]{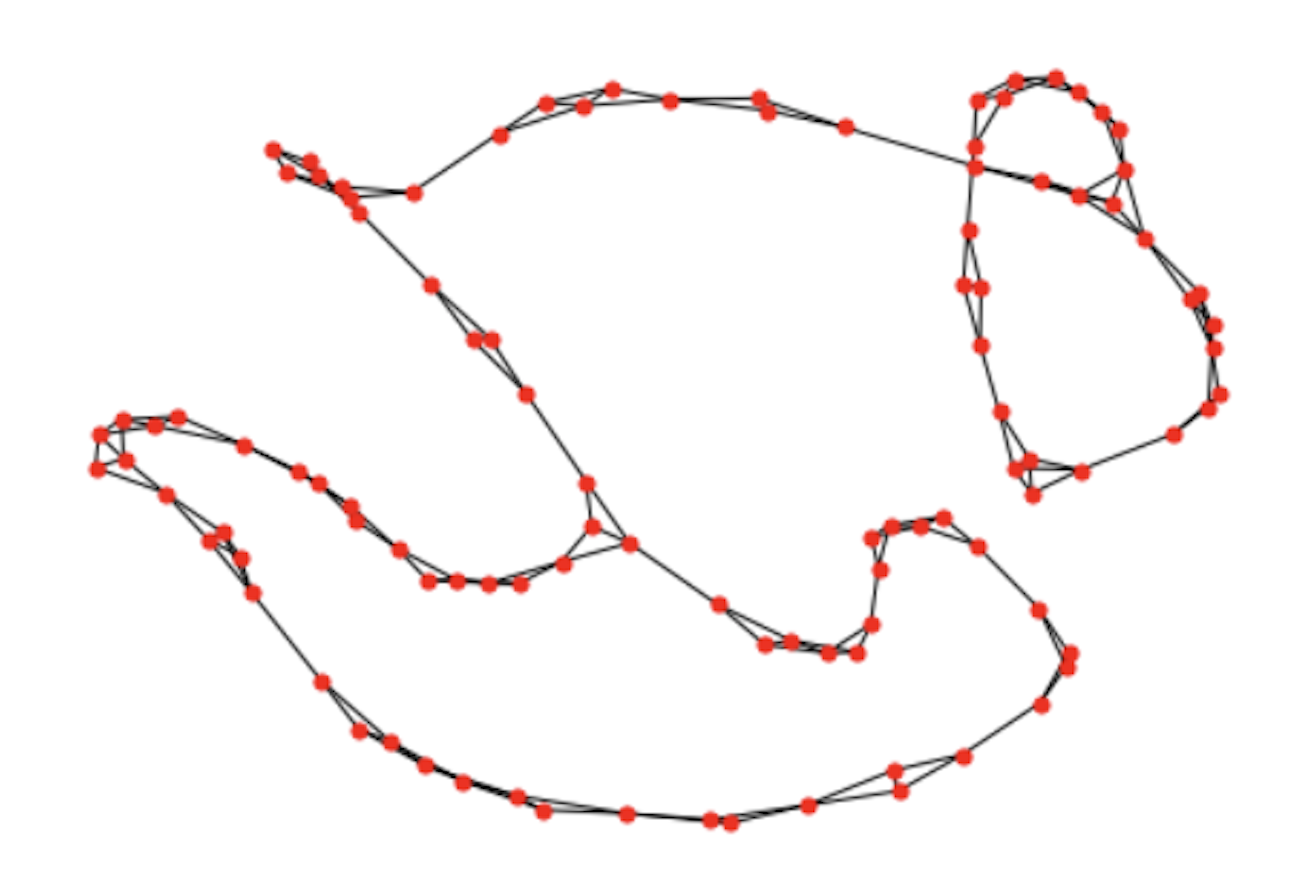} 
     \caption{Weighted Witness Graph}
\end{subfigure}%
\begin{subfigure}{0.25\linewidth}
    \centering \includegraphics[width=\linewidth]{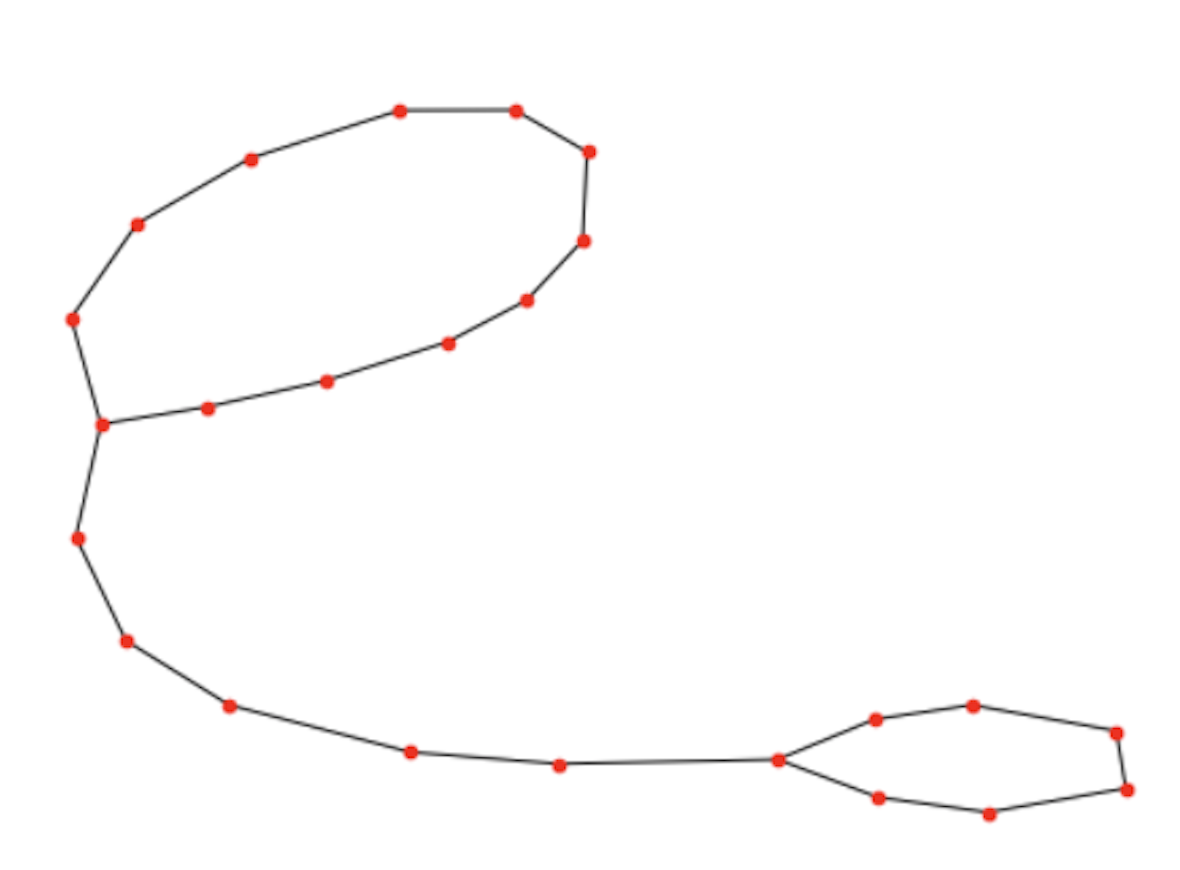} 
     \caption{ShapeVis}
\end{subfigure}
\caption{\footnotesize{An example pipeline of ShapeVis. (a): Original point-cloud data. (b): Sampled landmarks in the point cloud. (c): Weighted graph $G_L$ built on it. (d): Final graph generated by community detection and manifold tearing on $G_L$.}}
\end{figure*}\label{example_method}

\section{ShapeVis}
Given a large-scale and high-dimensional dataset $X = \{x_i \in \mathbb{R}^d, i=1,2,...,N\}$, our goal is to create a compressed graphical representation $G$ of $X$ while preserving the intrinsic structure of the data. Below, we explain the components of our visualization technique. 

\subsection{Manifold landmarking}
Our landmark selection procedure proceeds in two stages. In the first stage, for the given original input data $X \in \mathbb{R}^{d\times N}$ with $N$ records, points in $d$ dimensional space, we uniformly sample $M$ points, $X_M$. We then construct an undirected, unweighted neighborhood graph $G_M = (V_M, E_M)$, where each node $v_i \in V_M$, corresponding to the point $x_i \in X_M$, is connected to its $k$-nearest neighbors. Explicitly, an edge $e_{x_i,x_j} \in E_M$ if $x_j$ is in the $k$-nearest neighborhood set of $x_i$ or vice versa. This graph is further augmented using the remaining points in  $X \setminus X_M$ to build a $1$-witness complex. For a point $x_{r} \in X \setminus X_M$ let $x_p, x_q$ be its $2$-nearest neighbors from $X_M$. Then, we say the point $x_{r}$ is witnessing the $1$-simplex $\{x_p, x_q\}$ and we add an edge $e_{x_p,x_q}$ if not already present in the edge set $E_M$.  

In the second stage, we select the landmarks $L$ from $X_M$ using an inductive procedure similar to the one proposed in \cite{lisomap}. We start by selecting the first landmark $l_1$ from $X_M$ uniformly at random. At the $i$-th iteration, we mark the $k^{'}$-neighbors of the previously selected landmark $l_{i-1}$ as covered and remove them from $X_M$. These $k^{'}$-neighbors are termed as neighborhood set of landmark $l_{i-1}$.  We then inductively select another random point from the remaining set to be $l_i$ until all points in $X_M$ are marked. This algorithm ensures a selection of landmarks whose neighborhood-sets partition the graph.

\subsection{Random-walk based weighting} \label{l2landmarks}
Once we have sampled landmarks $L$ covering the underlying manifold, we construct a weighted, undirected graph $G_L$ on this set using the graph $G_M$ to capture its topology. Let $G_L = (V_L, E_L, W)$ where each node $v_i \in V_L$ corresponds to the landmark $l_i$. The edges $E_L$ and their weights $W$ are determined using a Finite Markov Chain to model the random movement of a hypothetical particle on the data manifold. The states are given by the landmarks. For each landmark $l_i$, we start $\beta$ random walks of fixed length $\theta_1 \leq \theta \leq \theta_2$ on $G_M$. We now define  
\begin{equation}
    a_{ij} = 
\begin{cases}
    \frac{n_{ij}}{ \sum_k n_{ik} },& \text{if } n_{ij}\geq th \\
    0,              & \text{otherwise}
\end{cases}
\end{equation}

where $n_{ij}$ denotes the number of random walks that started from landmark $L_i$ and have their endpoint in the neighborhood set of landmark $L_j$. This method generates a sparse asymmetric matrix $A = [a_{ij}]$ and the final weight matrix is given by the symmetric matrix $W = A+A^T - A\circ A^T$, where $\circ$ is the Hadamard (or pointwise) product. 

\begin{algorithm}[t]

\SetAlgoLined
\footnotesize{
    \textbf{Input}: $\{\textbf{X}\} \in \mathcal{D}_n; $\\
    \textbf{Output}: Graph $G$\\ 
    \Comment{Witness Complex creation}\\
    sample $X_{M} \subset X$ uniformly at random\\
    initialize $G_M=(V_M, E_M)$; the kNN graph on $X_{M}$ \\
    \For{$x \in \{X \setminus X_{M}\}$}
    {
        $x_p, x_q =   \text{NN}(x, X_{M}, 2)$; the $2$-nearest neighbors of $x$ in $X_M$ \\
        \uIf{ $e_{x_p, x_q} \not \in E_M$}
        { 
        $E_M = E_M \cup \{e_{x_p, x_q}\}$
        }
    }
    
    \Comment{Landmark selection}\\
    initialize $X_{L} := \varnothing $\\
    \While{$\text{len}(X_{M}) > 0$}
    {
        select $x$ from $X_{M}$ uniformly at random \\
        $X_{L} = X_L \cup \{x\} $ \\
        $X_{M} = X_M \setminus \{x \cup \text{Neigh}(x) \}$ \\
        set $\text{RevNeigh}(y) := x$ for each $y \in \text{Neigh}(x)$
    }
    \Comment{Weighted graph $G_{L}$ on $X_{L}$}\\
    initialize $N=[n_{ij}]=\textbf{0}$ \\
    \For{$\beta$ times}
    {
        random walk $p$ of length $\theta_1 \leq \theta \leq \theta_2$  \\
        let $l_i \in X_{L}$ and $l\in X_{M}$ be the starting and ending points of $p$ resp.\\
        let $\text{RevNeigh}(l) = l_j$ \\
        set $n_{ij} = n_{ij} + 1$ 
        
    }
    let $a_{ij} := n_{ij}/ \sum_k n_{ik}$\\
    let $w_{ij}:=a_{ij}+a_{ji}-a_{ij}\cdot a_{ji}$ \\
    let $G_L$ be the graph with adjacency matrix $W = [w_{ij}]$\\
    \Comment{Final visualization graph $G$ }\\
    Induced Graph $IG_p$ = CommunityDetection($G_{L}$)\\
    $G$ = ManifoldTearing($IG_p$) (as in Section \ref{manifold_Tearing})\\
    \textbf{return} $G= \{V,E\}$.
    }
 \caption{\footnotesize{ShapeVis}}
\label{ShapeVis_algo}
\end{algorithm}

\begin{figure*}[t]
\centering
\scalebox{0.9}{
\begin{subfigure}{0.25\linewidth}
    \centering \includegraphics[width=\linewidth]{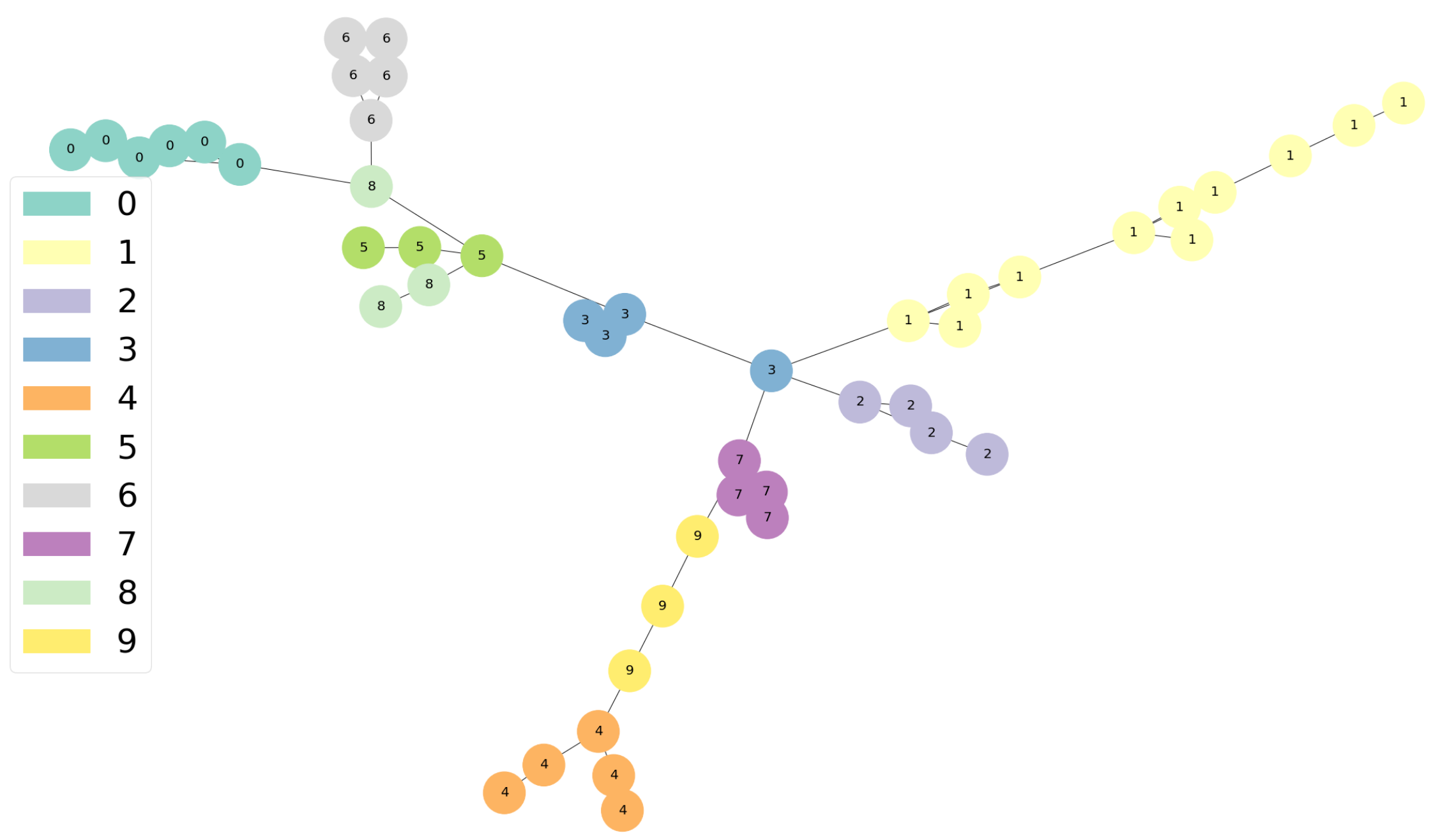} 
     \caption{ShapeVis}
\end{subfigure}%
\begin{subfigure}{0.25\linewidth}
    \centering \includegraphics[width=\linewidth]{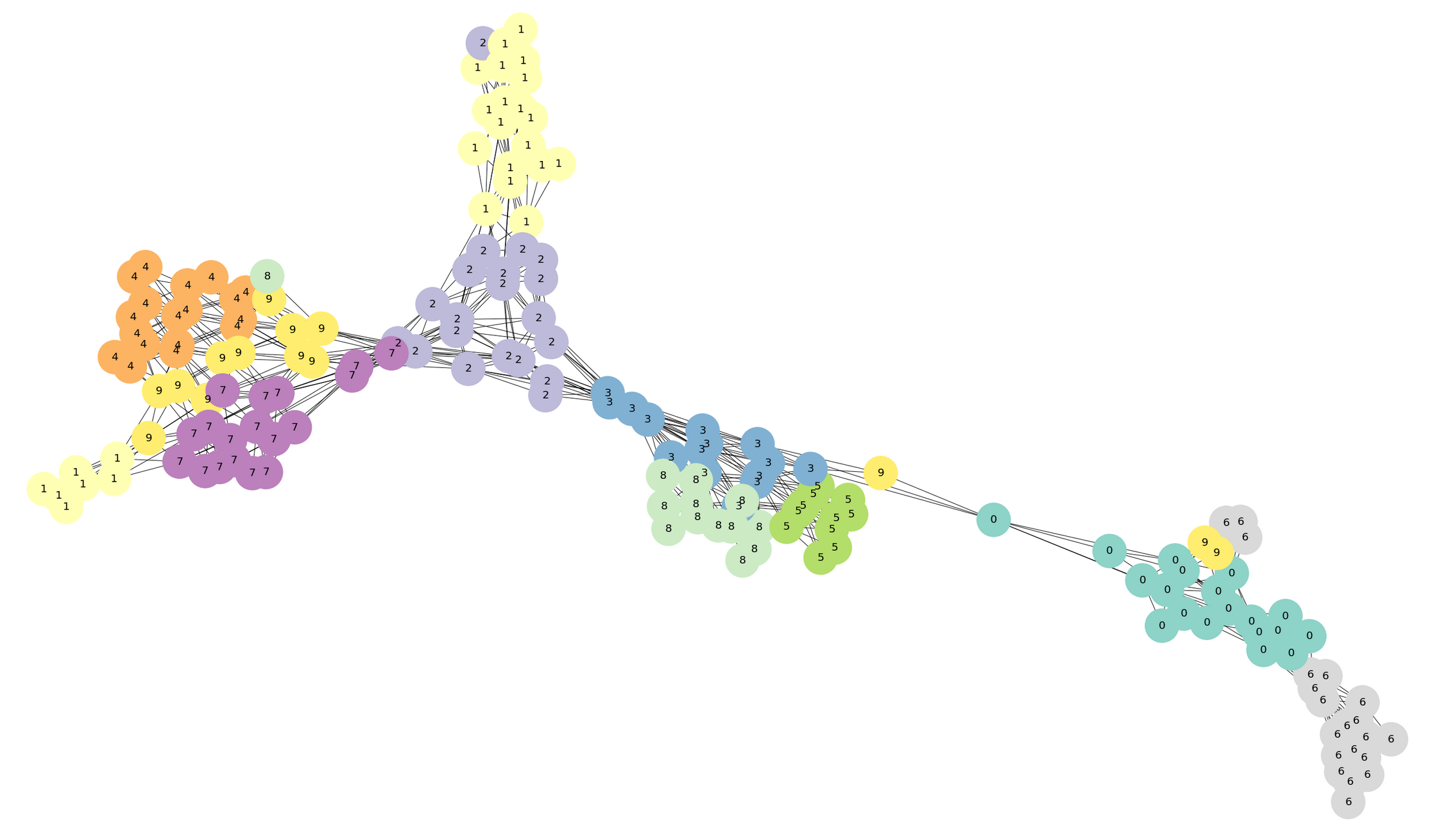} 
     \caption{Mapper(LargeVis)}
\end{subfigure}%
\begin{subfigure}{0.25\linewidth}
    \centering \includegraphics[width=\linewidth]{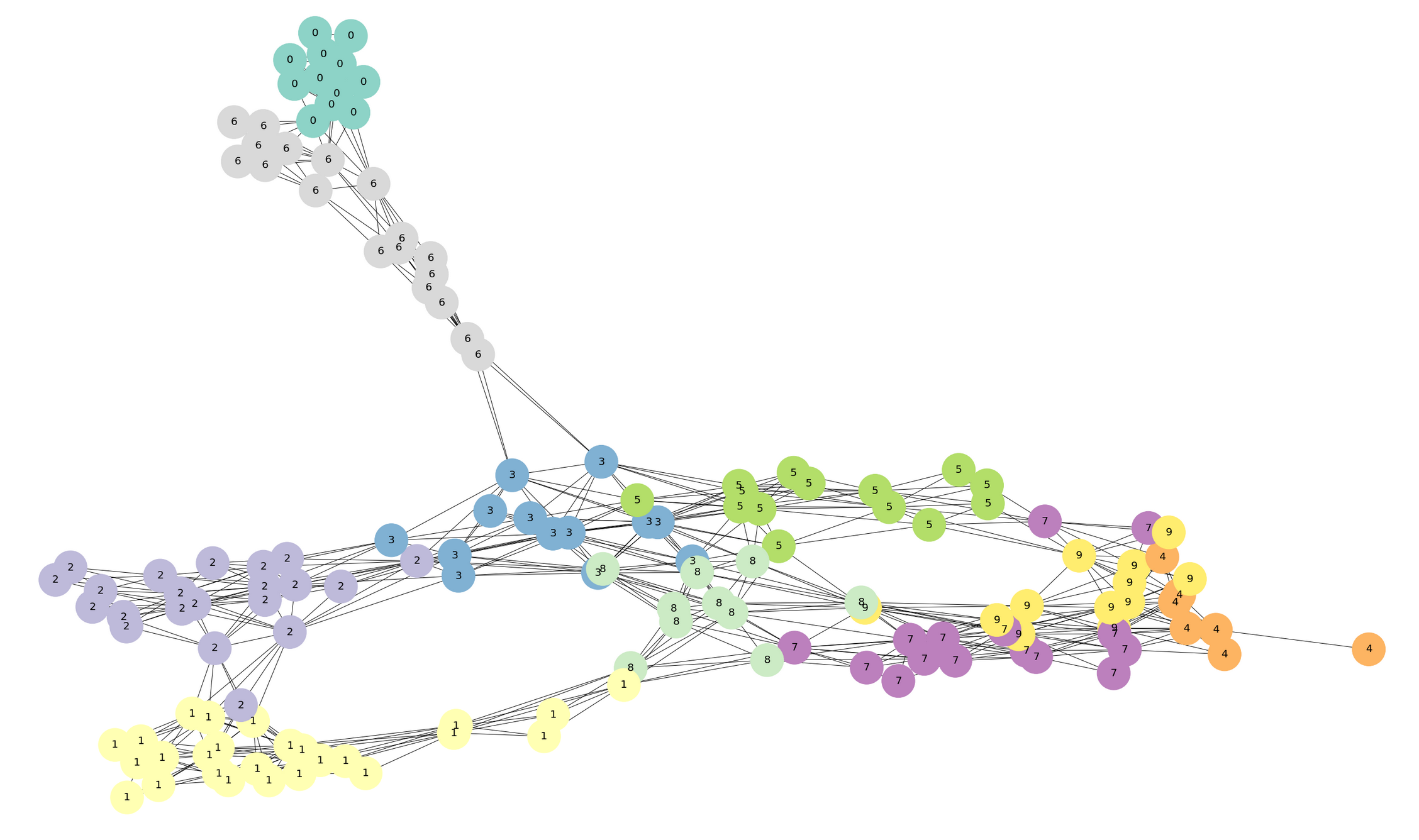} 
     \caption{Mapper(UMAP)}
\end{subfigure}%
\begin{subfigure}{0.25\linewidth}
    \centering \includegraphics[width=\linewidth]{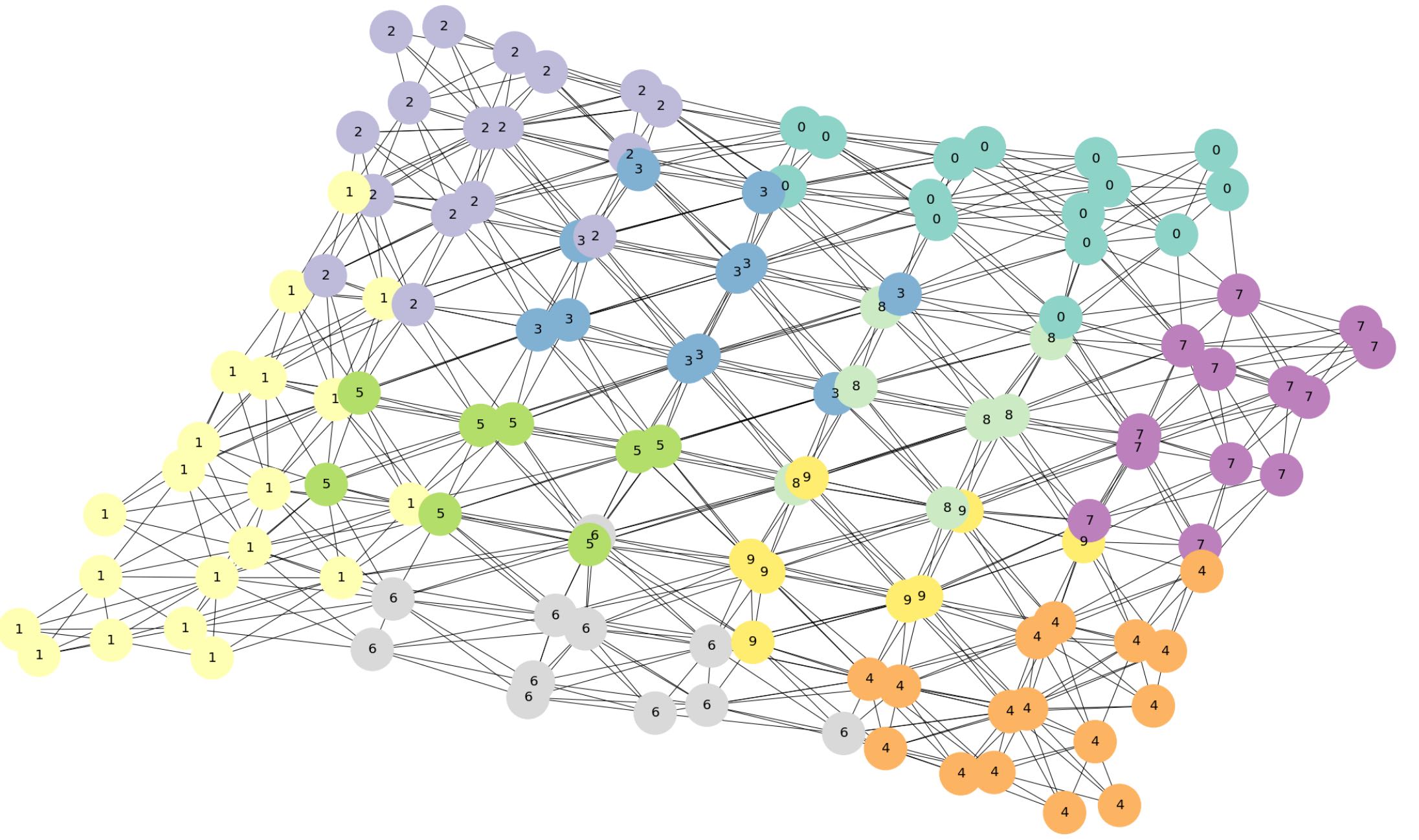} 
     \caption{Mapper(t-SNE)}
\end{subfigure}
}
\caption{\footnotesize{Visualization of MNIST dataset using different approaches}}
\label{mnist}
\end{figure*}

\begin{figure*}[t]
\centering
\scalebox{0.9}{
\begin{subfigure}{0.25\linewidth}
    \centering \includegraphics[width=\linewidth]{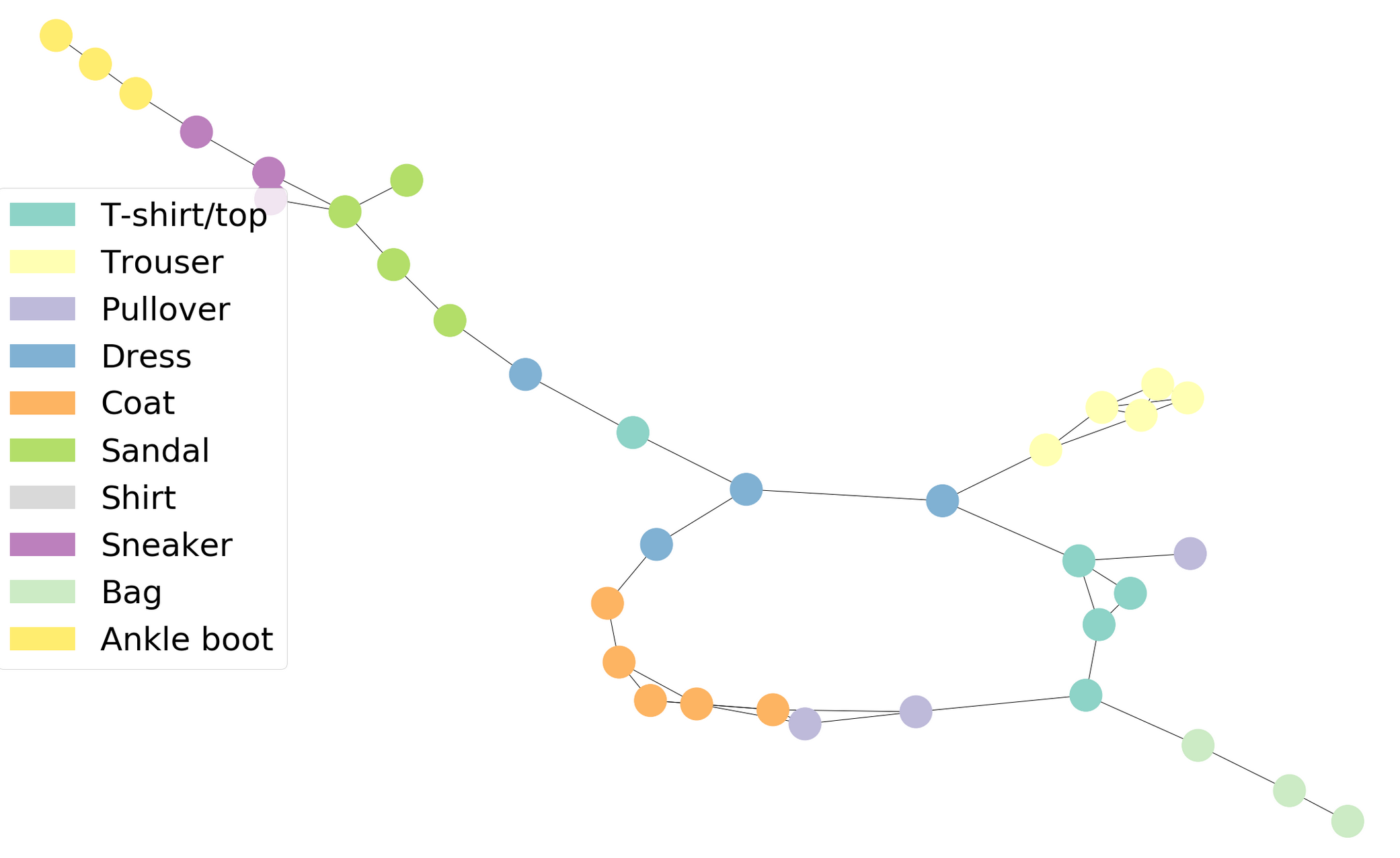} 
     \caption{ShapeVis}
\end{subfigure}%
\begin{subfigure}{0.25\linewidth}
    \centering \includegraphics[width=\linewidth]{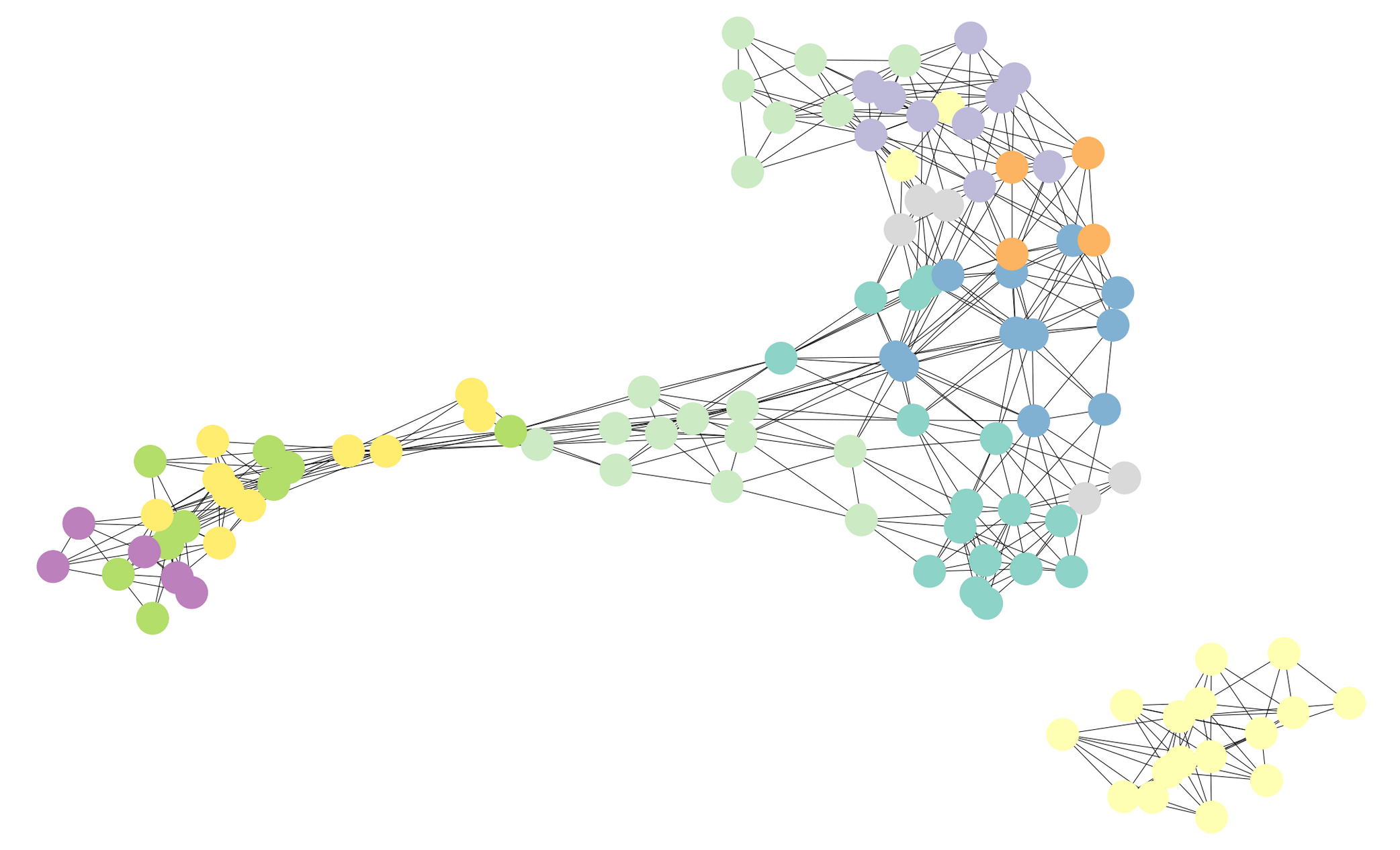} 
     \caption{Mapper(LargeVis)}
\end{subfigure}%
\begin{subfigure}{0.25\linewidth}
    \centering \includegraphics[width=\linewidth]{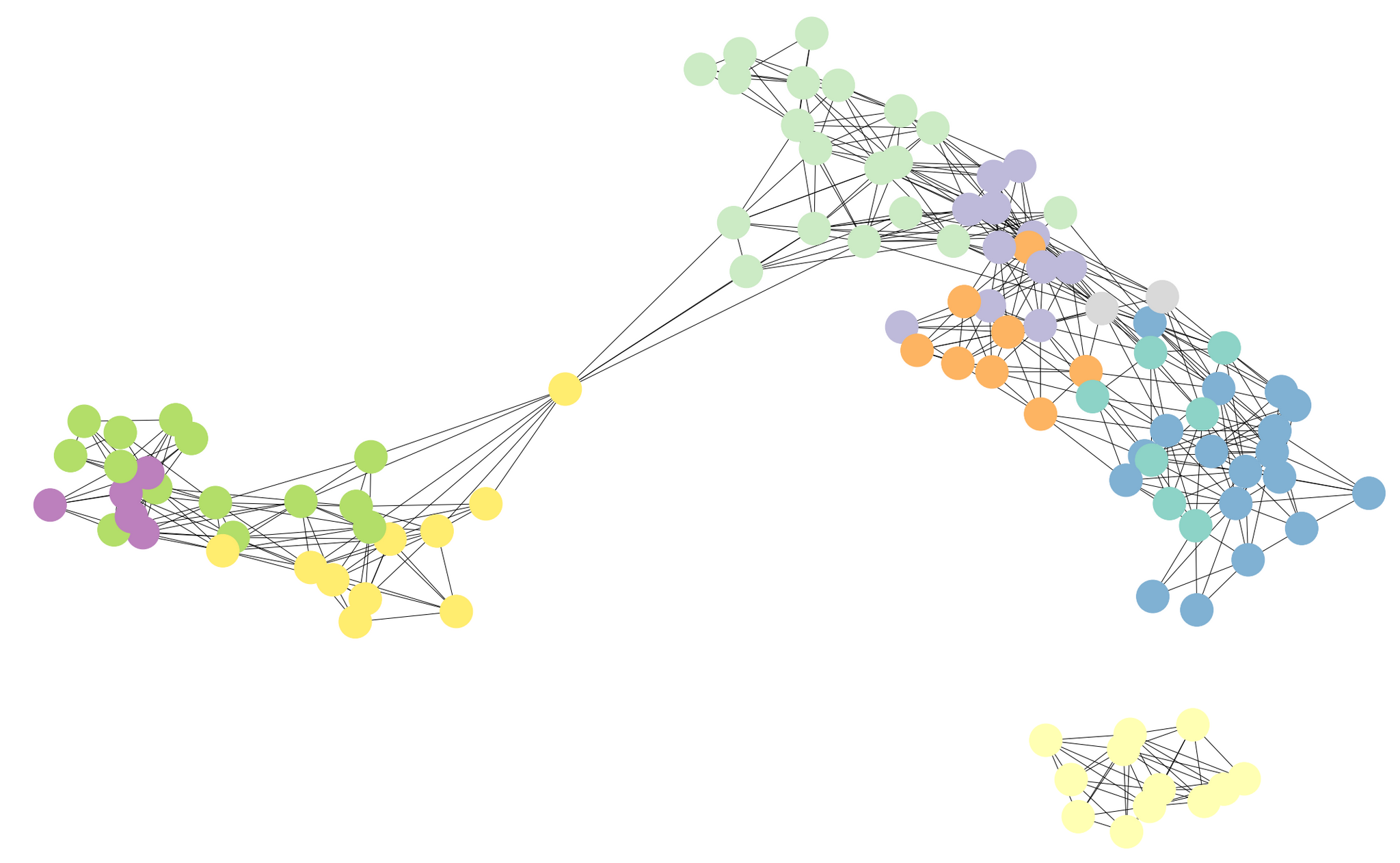} 
     \caption{Mapper(UMAP)}
\end{subfigure}%
\begin{subfigure}{0.25\linewidth}
    \centering \includegraphics[width=\linewidth]{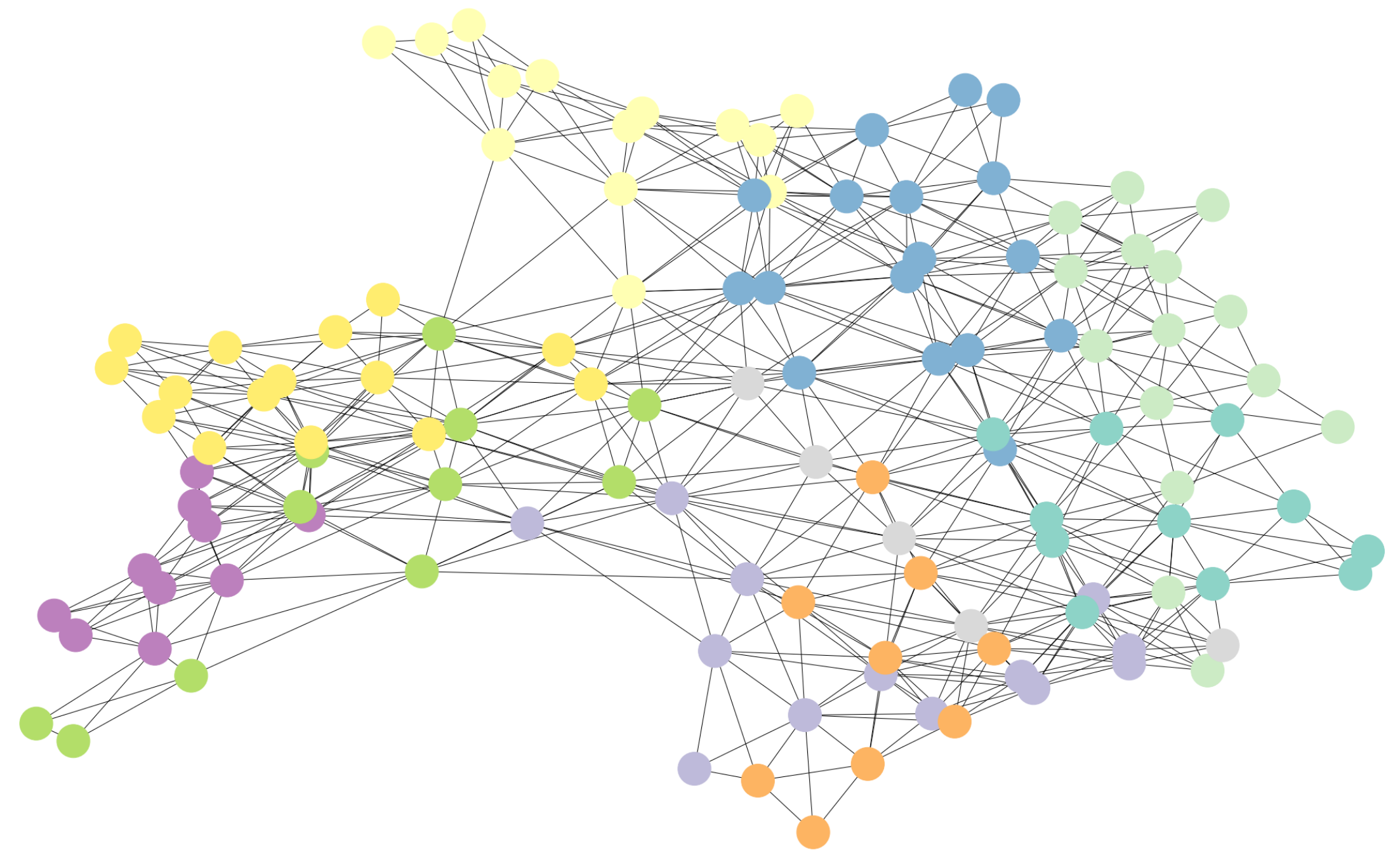} 
     \caption{Mapper(t-SNE)}
\end{subfigure}
}
\caption{\footnotesize{Visualization of FMNIST dataset using different approaches}}
\label{fmnist}
\end{figure*}

\subsection{Nerve complex of the graph}\label{nervecovering}

\textbf{Nerve of a cover}\cite{Hatcher}. An open cover of a space $X$ is a collection of open sets such that each point in the space is in at least one of these open sets. We shall refer to the individual elements of open cover as bins. We can conceptualize covering a space as putting each element in the space in one or more of these bins. Given a cover $\mathcal{U}$ of a space $X$, the nerve $N(\mathcal{U})$ is a simplicial complex constructed as follows: 
\begin{itemize}
    \item  The vertices (nodes) of $N(\mathcal{U})$ correspond to bins of $\mathcal{U}$
    \item If $k+1$ bins of $\mathcal{U}$ have a mutual non-empty intersection in $X$, $N(\mathcal{U})$ contains a $k$-simplex with the corresponding nodes as its vertices 
\end{itemize}

A topological space is said to be contractible if it is homotopy equivalent\cite{munkres2014topology} to a point. Basic examples of contractible spaces are the balls and more generally, the convex sets in $\mathbb{R}^d$. Open covers for which both elements and their intersections are contractible have the following remarkable property.


\begin{thm}[Nerve Theorem]\label{thm:nerve}\cite[Corollary 4G.3]{Hatcher} If $\mathcal{U}$ is an open cover of a paracompact space $X$ such that every non-empty intersection of finitely many sets in $\mathcal{U}$ is contractible, then $X$ is homotopy equivalent to the nerve $N(\mathcal{U})$.
\end{thm}

Graph $G_L$ obtained in Section \ref{l2landmarks} captures the shape of data, but we only want the higher-level homological features for insightful visualization. Nerve Theorem \cite[Corollary 4G.3]{Hatcher}, as explained above, provides a way to capture the topological structure of a space through a covering of the space. Inspired by this, we want to find a good covering of $X$ that captures its shape through the graph $G_L$. We use community detection algorithms \cite{louvain,leiden} to partition the graph into well-separated communities. An induced graph is constructed on this partition whose nodes represent the communities and an edge of weight $w$ exists between partitions if the sum of the weights of the edges between their elements is $w$. Using techniques from manifold tearing, we remove redundant/weak edges from the induced graph while preserving as much as possible the structural characteristics of the data manifold. 

\subsubsection{\textbf{Community Detection on Landmark graph}}

Community detection is performed on the graph $G_L$ obtained in Section \ref{l2landmarks} to obtain sets $\{C_i\}$ that cover the set $V_L$. This cover partitions the graph such that each node belongs to only one community. Since our visualization technique is unsupervised, we use modularity-based community-detection algorithms that use network structure properties, such as edges and node degrees to find communities. We can use any of the following standard algorithms for community detection: 

\paragraph{\textbf{Louvain}}\cite{louvain} This method uses a greedy optimization method that maximizes the modularity of a partition of the network. The optimization is done in two phases. First, individual nodes are moved to a neighboring community that yields the largest increase in modularity. Then an induced graph is created based on the partition obtained in the first phase. Each community in this partition then becomes a node in the induced network. These two phases are repeated until the modularity cannot be increased further.  

\paragraph{\textbf{Leiden}}\cite{leiden} This method is similar to Louvain except for a refinement phase. Here the optimization proceeds in three phases. First, nodes are moved based on modularity optimizations. In the refinement phase, nodes are not necessarily greedily merged with the community that yields the largest increase in modularity. Instead, a node may be merged with any community for which the modularity simply increases. The community with which a node is merged is selected randomly. The aggregation phase then proceeds similarly to the one in Louvain.

Both these methods give rise to a dendrogram structure where each level is a partition of the graph nodes. At the zeroth level is the first partition, which contains the smallest communities, and the final level contains the coarsest communities. The higher the level is, the bigger are the communities. Let the induced graph at partition level $p$ be $IG_p$.  

\vspace{-8pt}
\subsection{Modularity-based Manifold Tearing}\label{manifold_Tearing}

The induced graph $IG_p= (V_p; E_p)$ obtained in Section \ref{nervecovering} is dense with spurious edges and hence does not lead to a comprehensible representation of the data. This step aims at determining a graph $G = (V; E)$, having the same vertices as the graph $IG_p$, but with a smaller edge set $E$, such that $G$ represents the overall topological structure of $G_L$. We introduce a two-phase tearing procedure to construct $G$. 

In the first phase we construct a spanning sub-graph $G_S$ on $IG_p$. For each edge of the graph $IG_p$, its modularity is computed and inserted in an ordered heap of edges. We then iteratively pop from the heap and introduce the corresponding edge into $G_S$ if it results in increased connectivity of the graph until the graph has as many connected components as $IG_p$. This phase results in a sub-graph $G_S = (V; E_1) $ that spans the vertices of the induced graph. This procedure differs significantly from the classical manifold tearing one in the following sense. Instead of constructing a minimum spanning tree (MST) or a shortest path tree (SPT) with no cycles on the graph; our procedure constructs a spanning subgraph whose edges are chosen based on their modularity. 
\begin{table*}[t]
  \centering
\scalebox{0.9}{
  \begin{tabular}{ | c | c | c | c | }
    \hline
    Dataset & ShapeVis & Mapper(UMAP) & Mapper(LargeVis) \\ \hline
    \begin{tabular}{@{}c@{}c@{}c@{}}LiveJournal \\ \small(pseudo labels based on \\
    \small segments in \\
    \small ShapeVis)\end{tabular}  & 
    \begin{minipage}{.25\textwidth}
      \includegraphics[width=\linewidth]{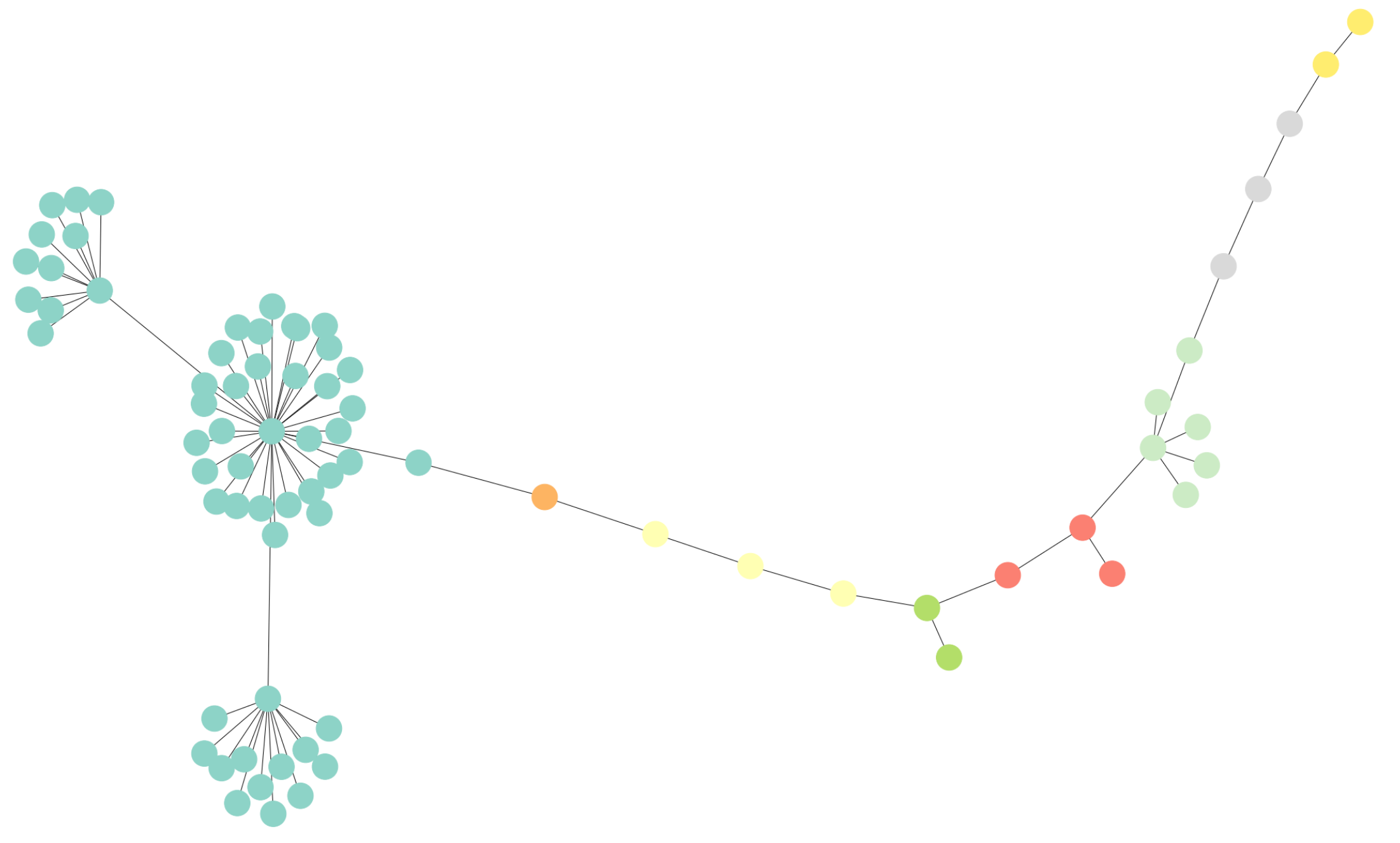}
    \end{minipage}
    &
    \begin{minipage}{.25\textwidth}
      \includegraphics[width=\linewidth]{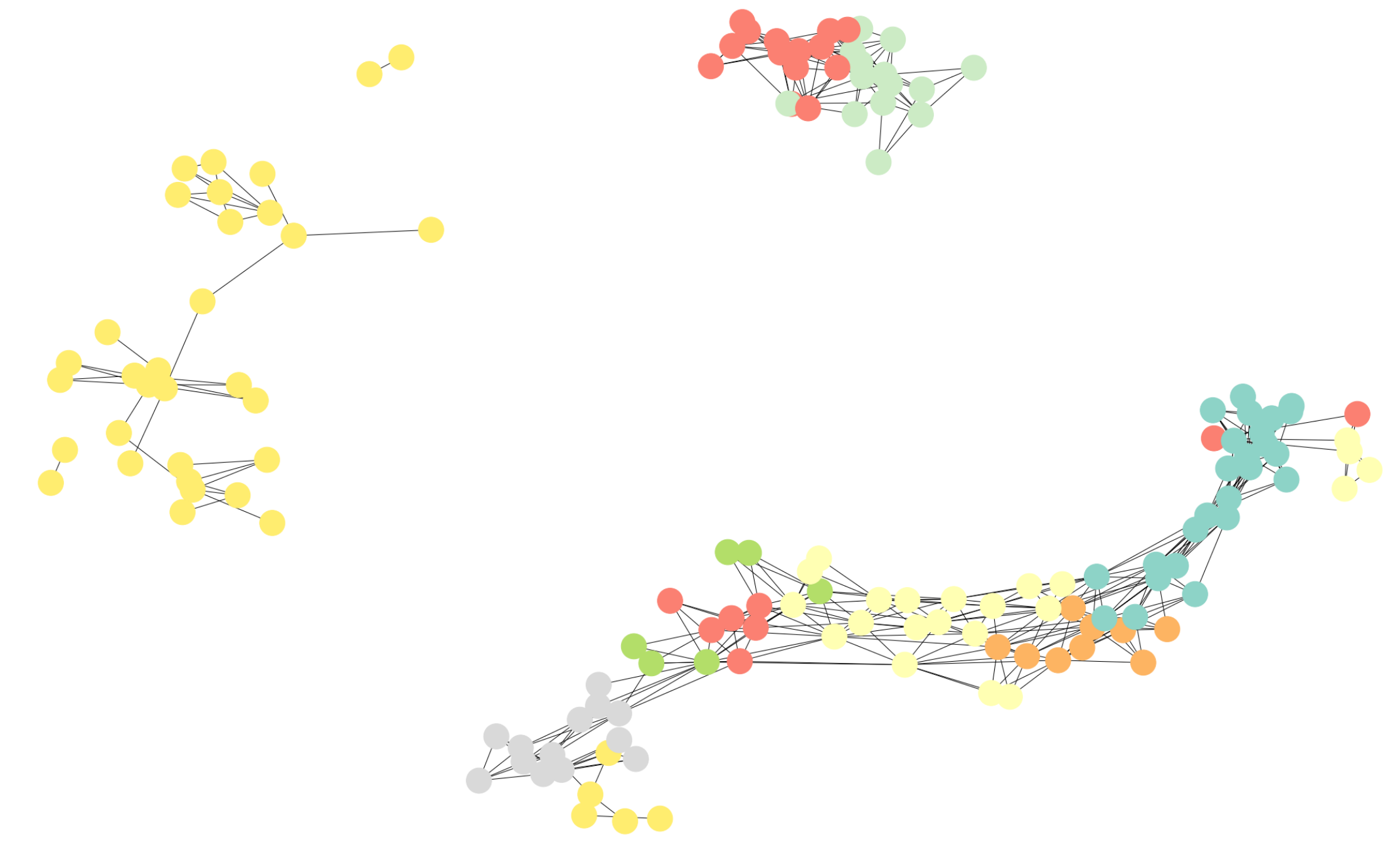}
    \end{minipage}
    &
    \begin{minipage}{.25\textwidth}
      \includegraphics[width=\linewidth]{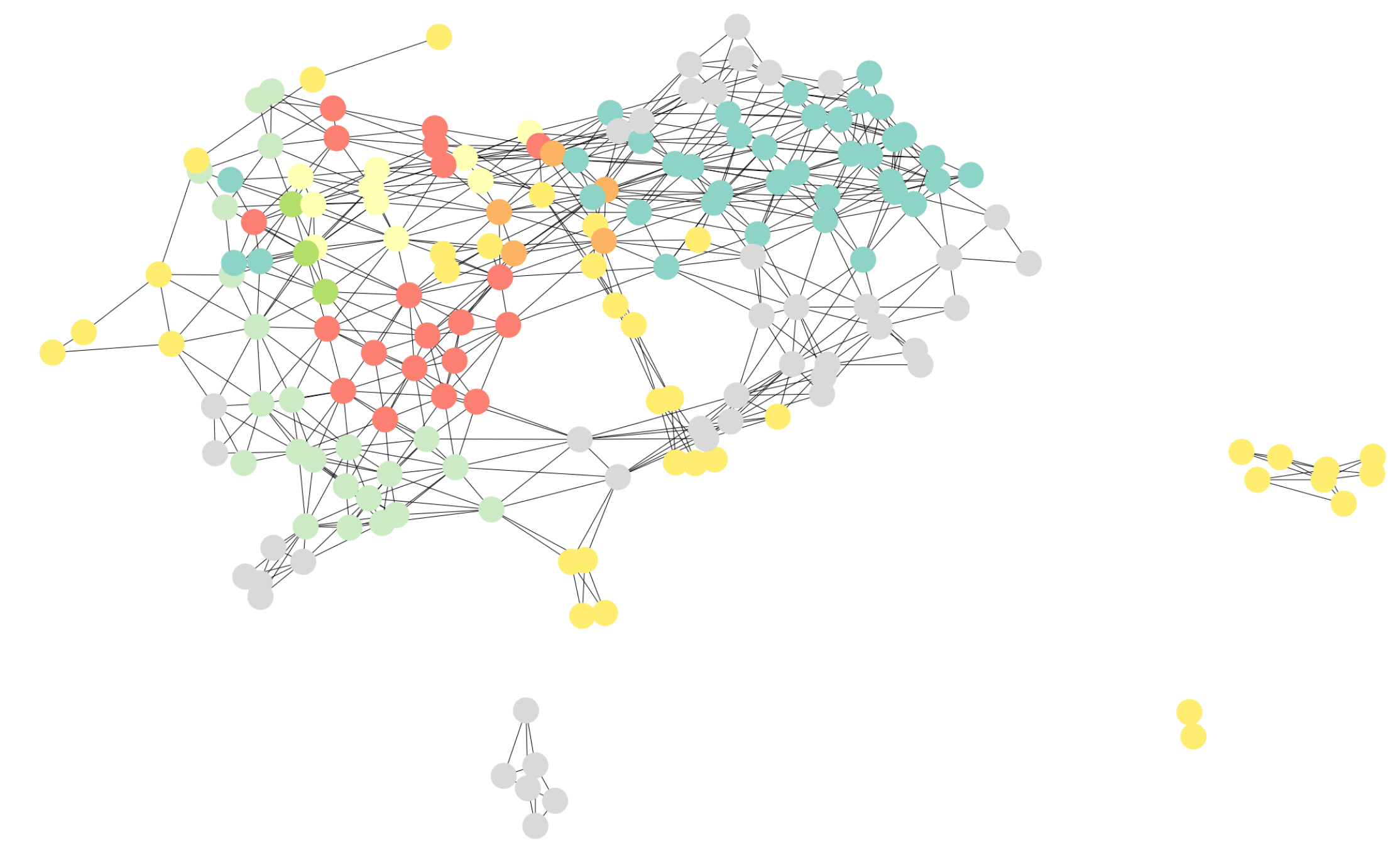}
    \end{minipage}
    \\ \hline
    \begin{tabular}{@{}c@{}c@{}c@{}}LiveJournal \\ \small(pseudo labels based on \\
    \small segments in \\
    \small Mapper(UMAP))\end{tabular}  & 
    \begin{minipage}{.25\textwidth}
      \includegraphics[width=\linewidth]{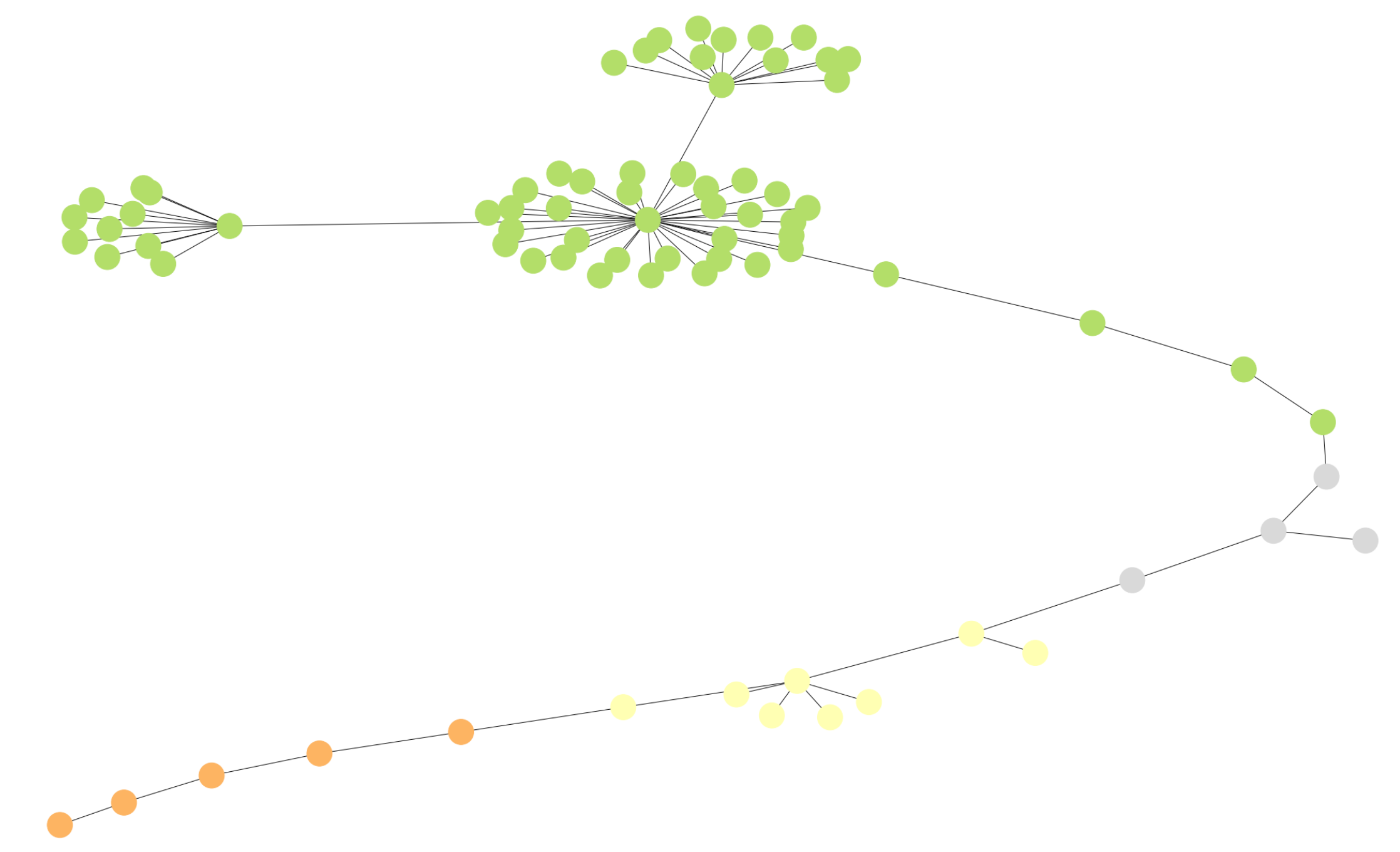}
    \end{minipage}
    &
    \begin{minipage}{.25\textwidth}
      \includegraphics[width=\linewidth]{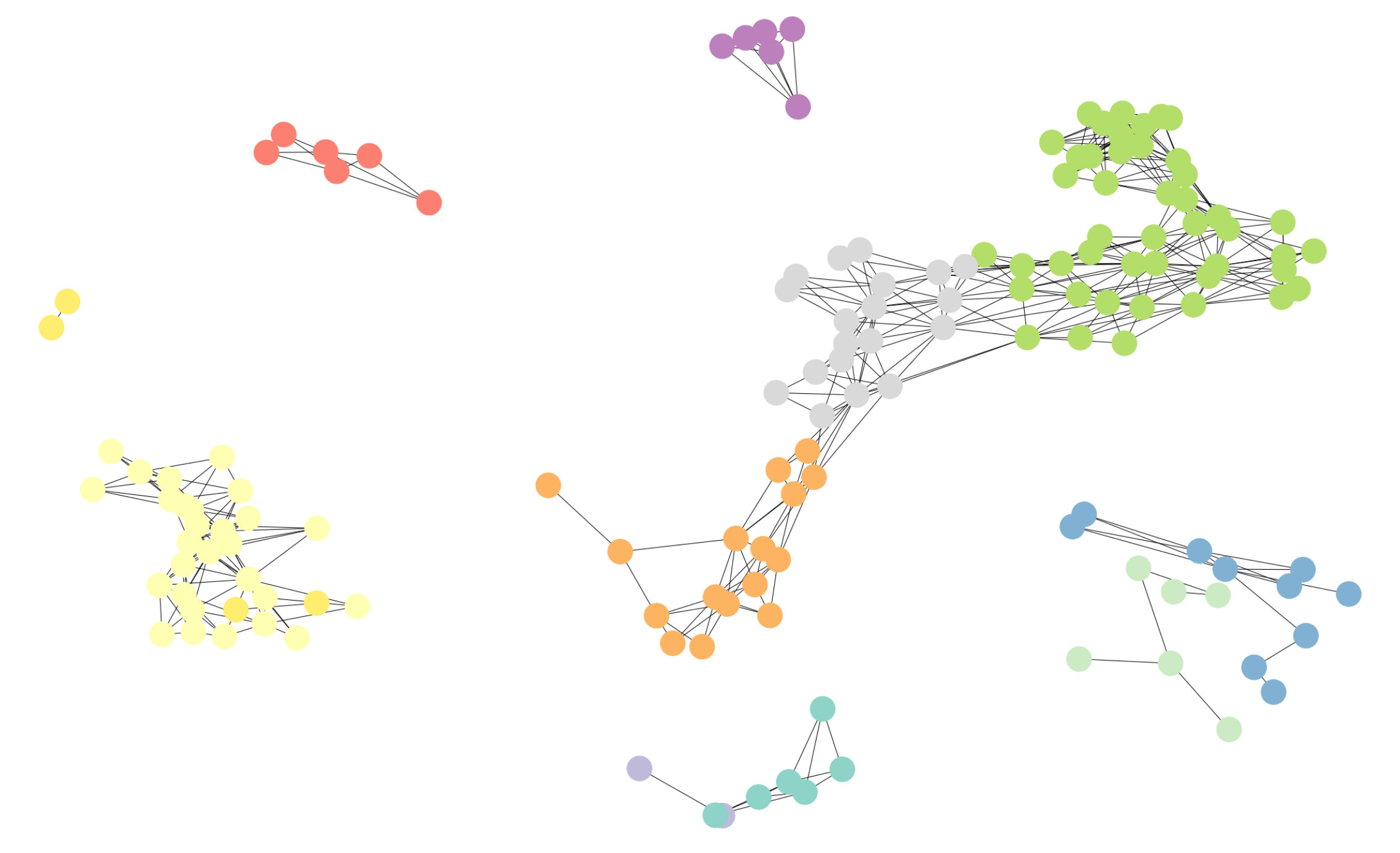}
    \end{minipage}
    &
    \begin{minipage}{.25\textwidth}
      \includegraphics[width=\linewidth]{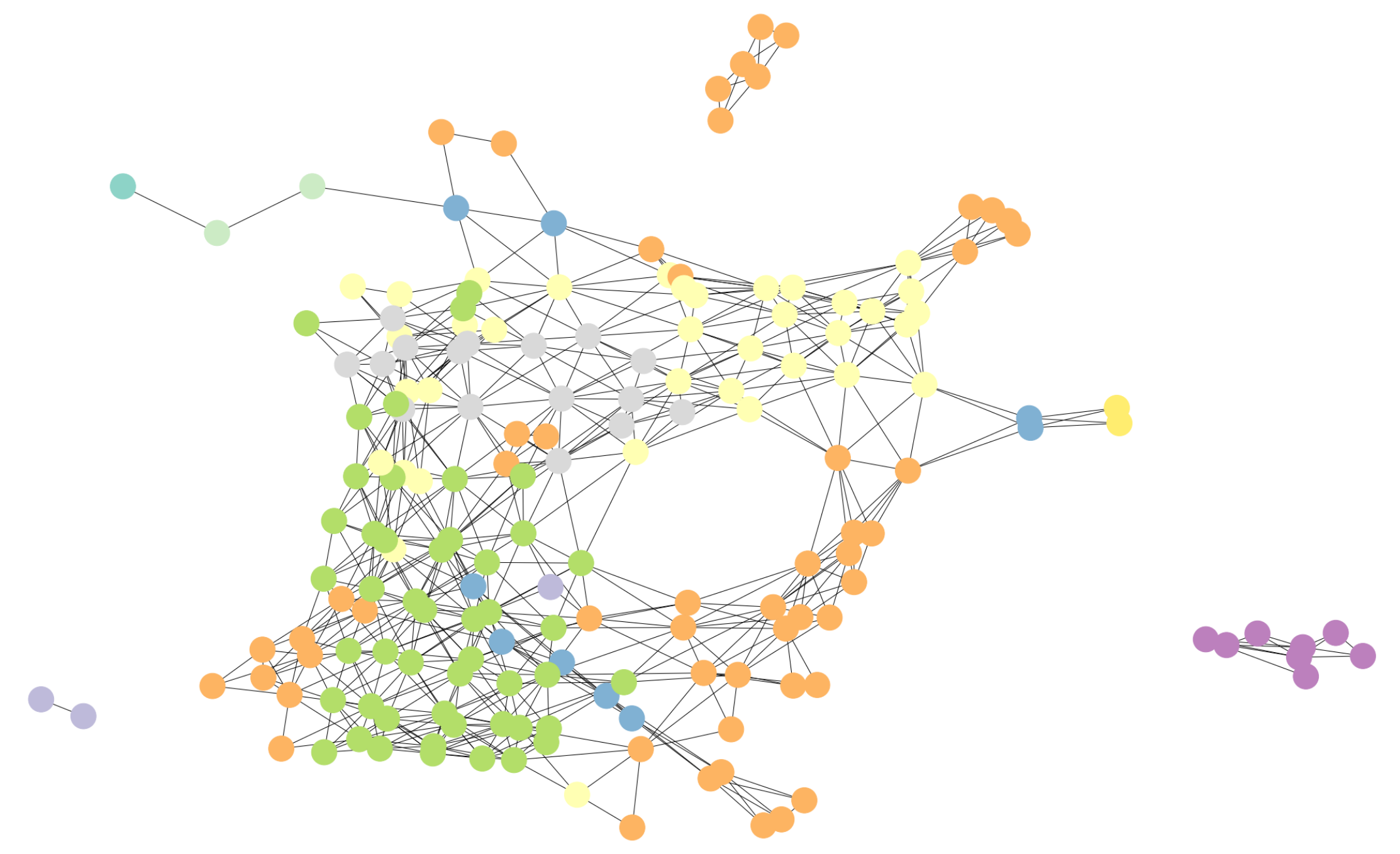}
    \end{minipage}
    \\ \hline
    \begin{tabular}{@{}c@{}c@{}c@{}c@{}}GoogleNews \\
    word vectors\\
    \small(pseudo labels based on \\
    \small segments in \\
    \small ShapeVis)\end{tabular}  & 
    \begin{minipage}{.25\textwidth}
      \includegraphics[width=\linewidth]{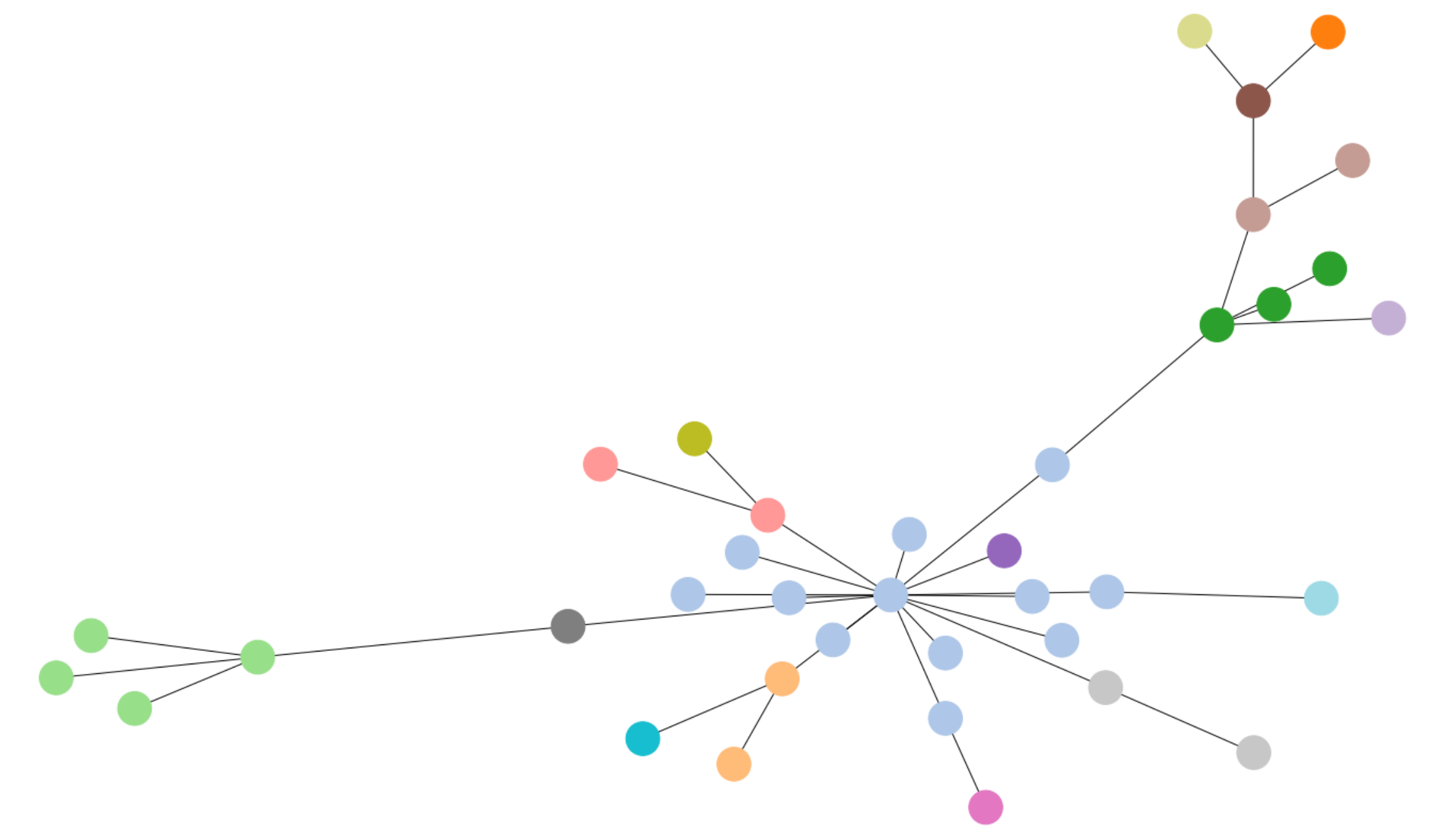}
    \end{minipage}
    &
    \begin{minipage}{.25\textwidth}
      \includegraphics[width=\linewidth]{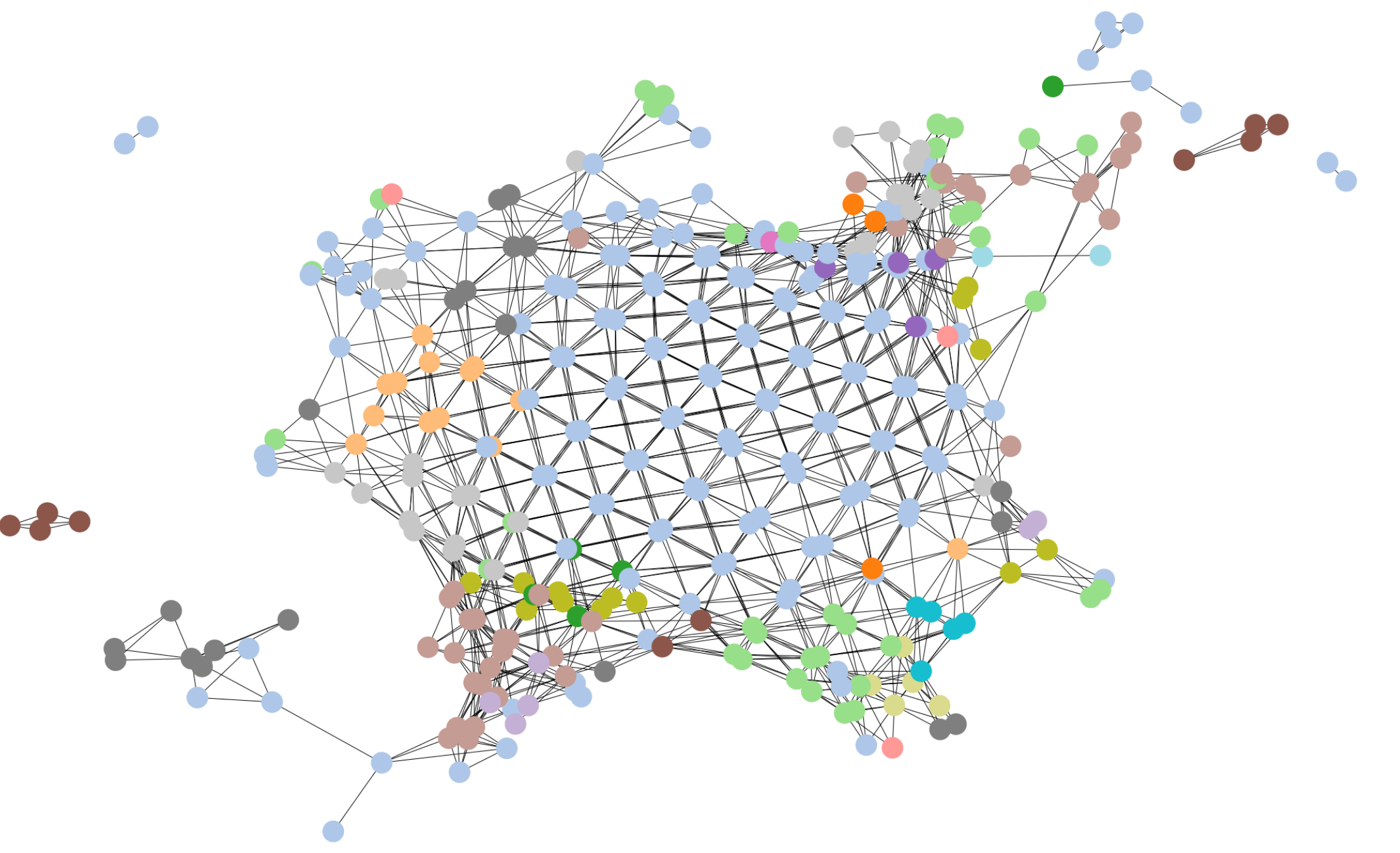}
    \end{minipage}
    &
    \begin{minipage}{.25\textwidth}
      \includegraphics[width=\linewidth]{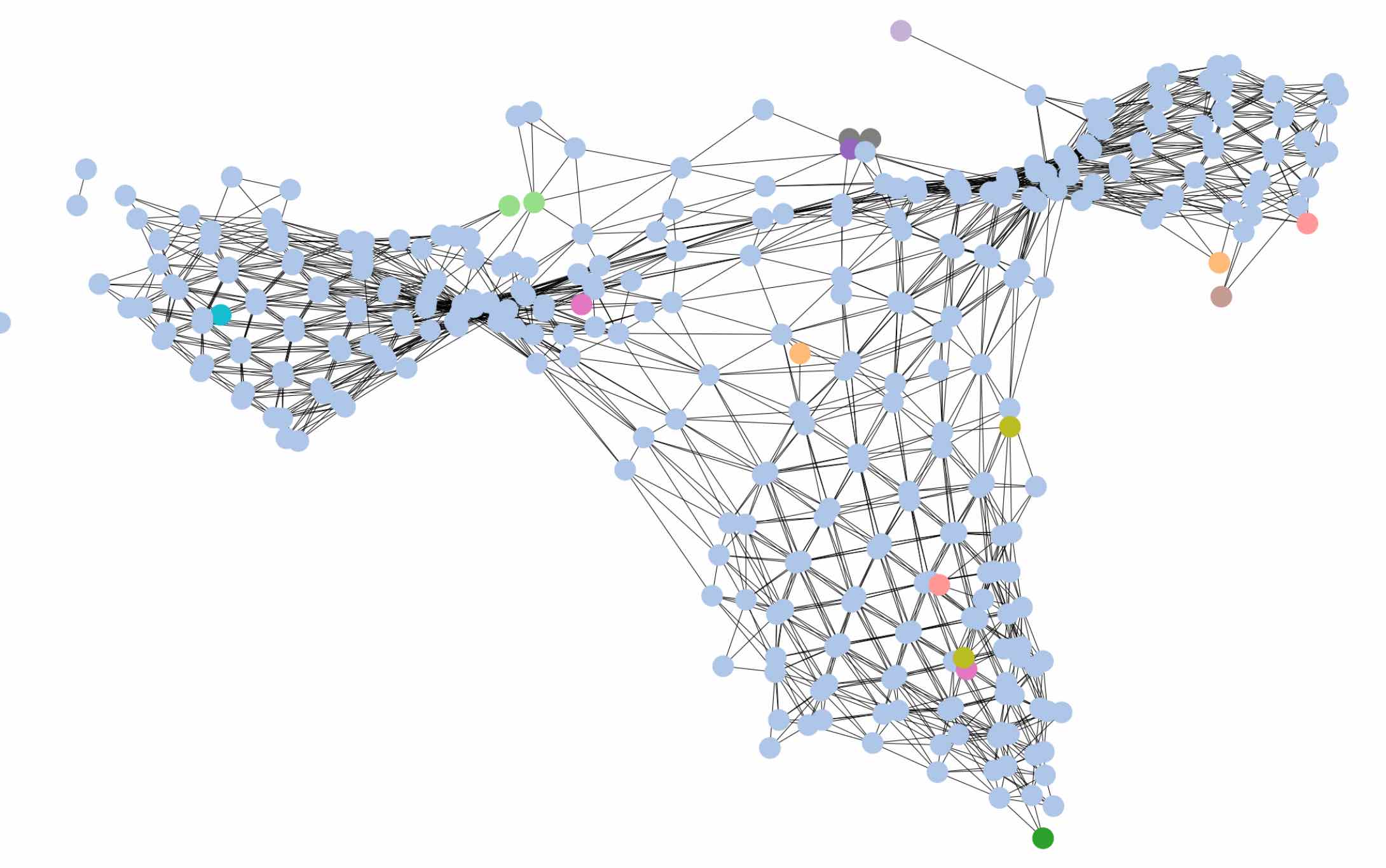}
    \end{minipage}
    \\ \hline
    \begin{tabular}{@{}c@{}c@{}c@{}c@{}}GoogleNews \\
    word vectors\\
    \small(pseudo labels based on \\
    \small segments in \\
    \small Mapper(UMAP))\end{tabular}  & 
    \begin{minipage}{.25\textwidth}
      \includegraphics[width=\linewidth]{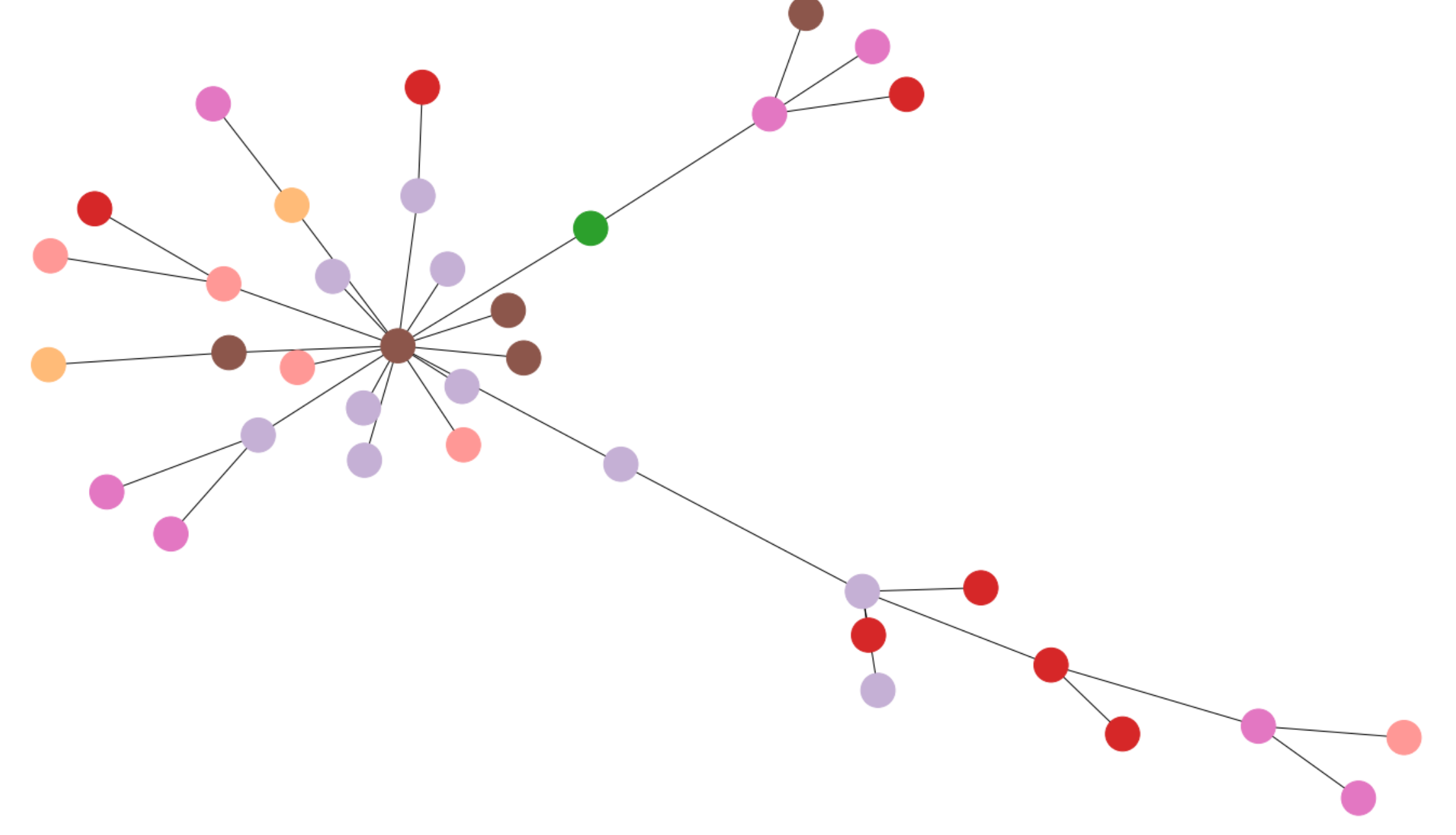}
    \end{minipage}
    &
    \begin{minipage}{.25\textwidth}
      \includegraphics[width=\linewidth]{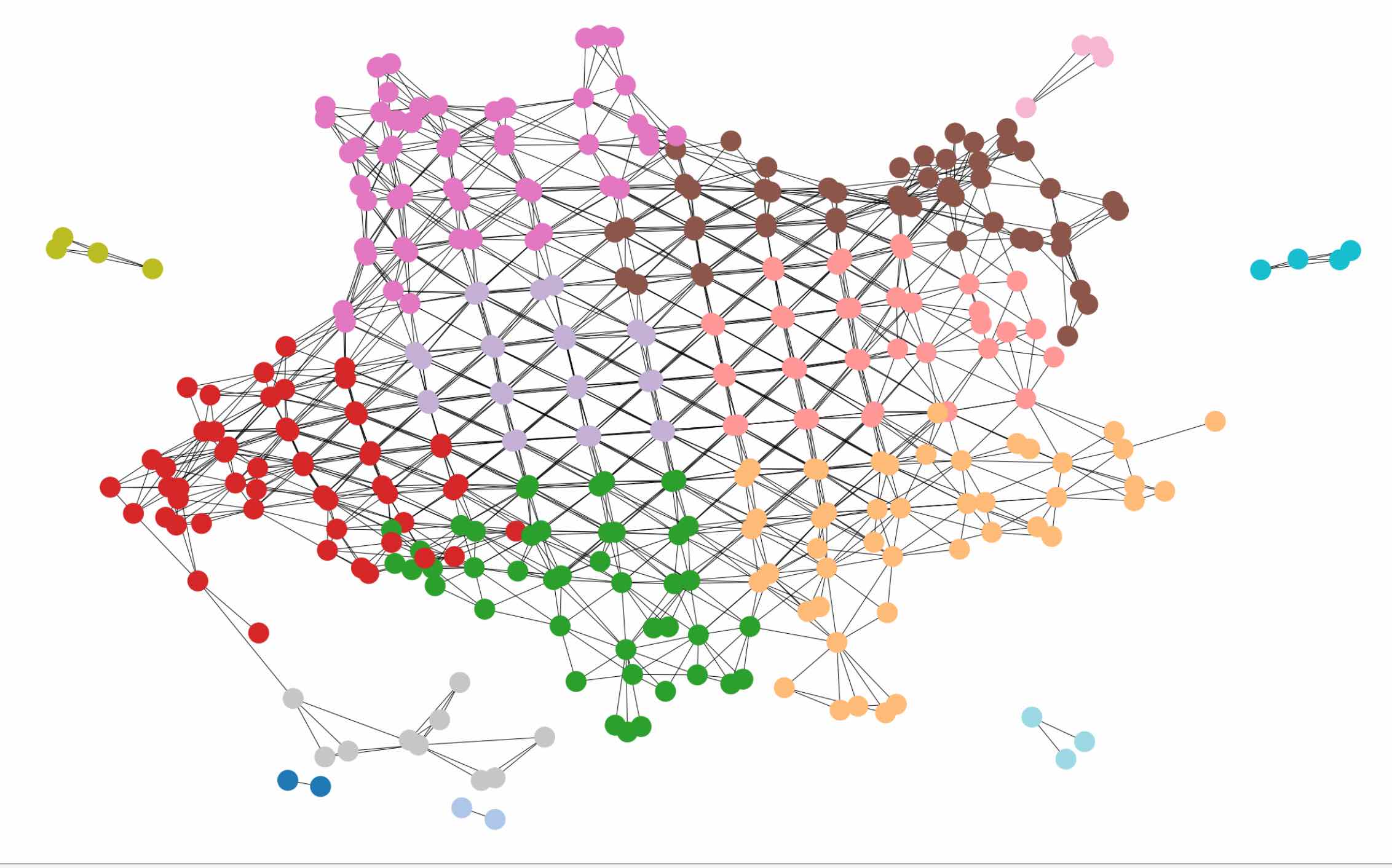}
    \end{minipage}
    &
    \begin{minipage}{.25\textwidth}
      \includegraphics[width=\linewidth]{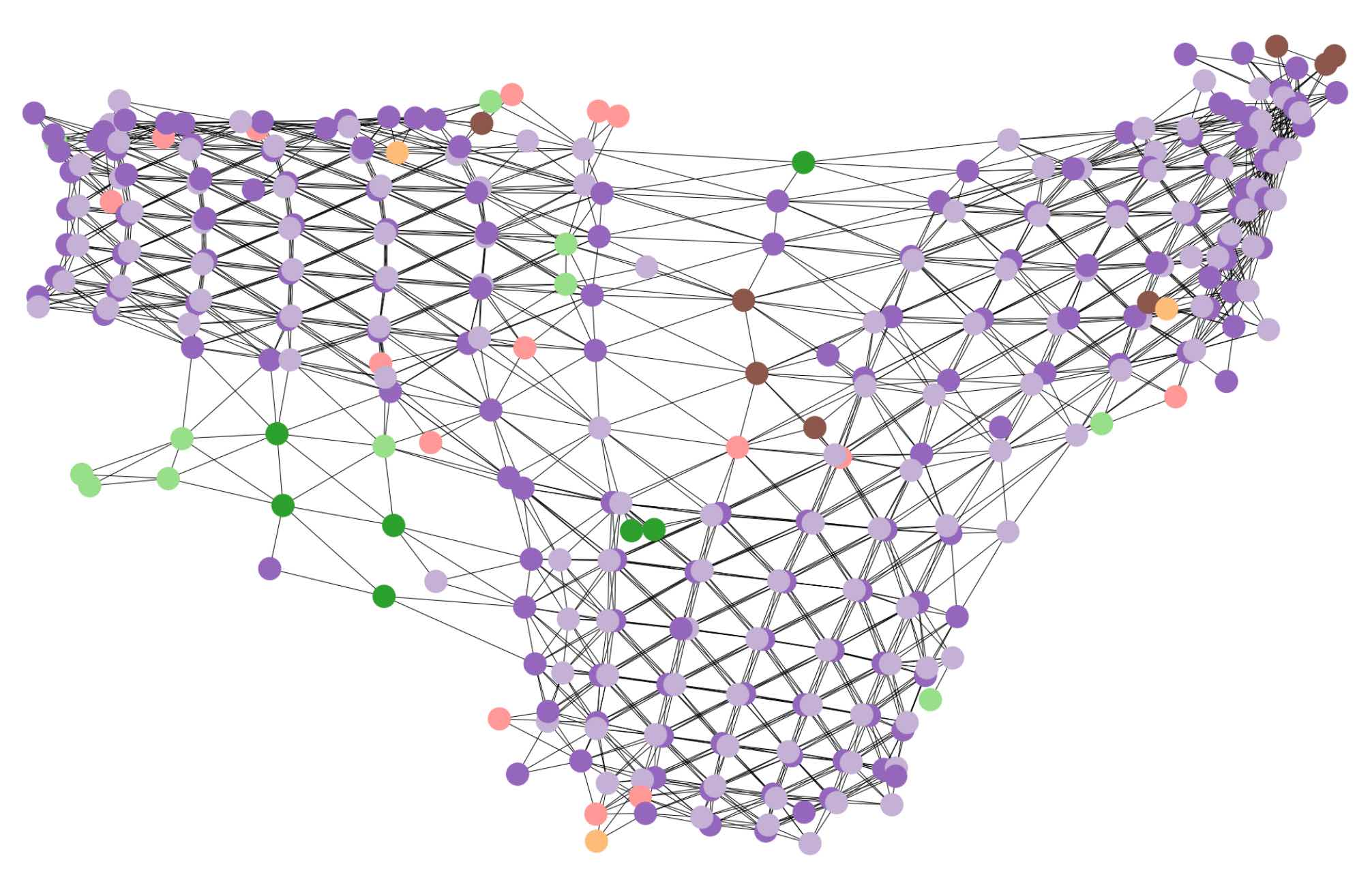}
    \end{minipage}
    \\ \hline
    
  \end{tabular}
}
  \captionof{figure}{\footnotesize{Visualization of LiveJournal (Rows 1 and 2) and GoogleNews word vectors (Rows 3 and 4) using different approaches. Nodes are colored by the pseudo label whose data points are maximum in the node.}}\label{lj_and_w2v}
\end{table*}

Once the spanning subgraph $G_S$ is computed, we execute the second phase of the tearing procedure. Whereas classical techniques cut out essential loops to aid in downstream dimensionality-reduction, we introduce as few loops as possible to capture the structure of the data manifold as much as possible. We initialize $G$ with the spanning subgraph $G_S$ and gather the edges discarded during the first phase in a set $S = E_p\setminus E_1$. In this phase, our procedure only reintroduces those edges from $S$ that generate essential loops. By essential loops we mean those cycles whose sum of edge modularities is more than or equal to $c$, a user-defined hyperparameter. The idea is to preserve the homological characteristics of the data manifold like connected components and loops. 

Thus, the final output is the graph $G=(V,E)$  where local coverings of data are represented as nodes and edges represent the geodesic proximity between them. Branches and loops reflect the topological features of data manifold and are useful in the interactive and unsupervised discovery of segments.
In the next section, we show the visualization graph obtained through our approach on high-dimensional and large scale datasets. An example pipeline on synthetic data is shown in Fig 1. 

\section{Experiments}
In this section, we evaluate the effectiveness of our approach in visualizing high-dimensional and large-scale datasets. We compare our visualization approach with the following approaches:
\begin{itemize}
    \item Mapper(t-SNE): Mapper with t-SNE filter function. 
    \item Mapper(UMAP): Mapper with UMAP filter function. 
    \item Mapper(LargeVis): Mapper with LargeVis filter function. 
\end{itemize}

We choose these baselines as Mapper also returns a compressed visualization of data in the form of a graph similar to ShapeVis. Dimensionality reduction algorithms are a standard choice of filter function in Mapper to compress the data into a 2-dim space, and t-SNE\cite{tsne}, UMAP\cite{umap}, LargeVis\cite{largevis} are widely used state-of-the-art dimensionality reduction algorithms. More details about Mapper algorithm can be found in \cite{mapper}. In the next section, we show that visualization quality of our approach is comparable and sometimes on-par to the above-mentioned approaches while being scalable to millions of data-points.

\subsection{Datasets}
We use the following datasets for comparison and visualization purposes. It includes both high-dimensional and large-scale real-world datasets. 

\begin{itemize}
    \item MNIST \cite{mnist_dataset}. The dataset consists of 70000 28x28images of handwritten digits (0-9). Each data-point is a 784-dimensional vector.
    \item F-MNIST \cite{fmnist}. Fashion MNIST is a dataset of 28x28images of fashion items like clothing, shoes etc. There are 10 classes and total of 70000 images. 
    \item GoogleNews Vectors \cite{word2vec}. It is a dataset of 3 million words and phrases from GoogleNews dataset. Each word is embedded into a 300-dimensional vector space using word2vec \cite{word2vec} approach. 
    \item LiveJournal \cite{livejournal}. It is a social network dataset from an online blogging community with around 4 million nodes. For common comparison with other methods, we first learn a 100-dimensional representation of each node using LINE \cite{line} algorithm before visualizing it. Note that our algorithm can be easily modified to work directly over graphs as well. 
\end{itemize}



\subsection{Implementation Details}\label{appendix}
We explain our choice of hyperparameters for ShapeVis implementation. We initialize $M$, which is the number of points sampled for creating witness complex $G_M$, as a minimum of 1 Million or $N/3$. $N$ is the total number of data points. For the construction of k-nn graph $G_M$ we keep the number of nearest neighbours $k$ fixed at $10$ for all datasets. Note that $k$ should be such that $G_M$ captures the local connectivity in the dataset and therefore a small value of $k$ is good if the sample size is large enough. For finding k-nearest neighbors we use the nn-descent algorithm of \cite{nndescent}. Number of random walks $\beta$ is $1000$ and $\theta_1$, $\theta_2$ is fixed as $l/2$, $l$ where $l=50$ for all datasets. We found the algorithm to be stable for various choices of $l$ and $\beta$ greater than a threshold and chose the minimum to optimize time. For $IG_p$ we choose partition level $p$ as $0$ and find it to work best for all datasets. Similarly, the parameter $c$ during manifold tearing step is kept fixed at $2*log$(Modularity of $IG_p$). 

For Mapper, we show the best visualization obtained by tuning hyper-parameters wherever possible. For LargeVis and UMAP we use the reference implementation of \cite{largevis} and \cite{umap}, and for t-SNE, we use Multicore t-SNE package \cite{multicore-tsne}.

\subsection{Qualitative Comparison}
\paragraph{\textbf{Labeled datasets}} For comparing visualization quality on datasets with ground truth labels, we color each node in the visualization graph with its dominant label. Fig \ref{mnist} shows the visualization obtained on MNIST dataset. We can see that ShapeVis as well as Mapper(UMAP) and Mapper(LargeVis) coherently capture the global relationship between different digits with clusters of (1,2), (3,5,8), (0,6) and (4,7,9). But ShapeVis also captures the local within-cluster relationships, e.g. the two branches of '4' in Fig \ref{mnist} (a) corresponds to the two different ways of writing it: upside-down lower-case 'h' and closed digit '4'. Fig \ref{fmnist} shows the visualization obtained on F-MNIST dataset. Though all the visualizations show the separation between the two broad category clothing and footwear, ShapeVis captures the relationship between different classes more coherently. For example, \textit{Trouser} class is connected to \textit{Dress} class through a single node instead of being disconnected with the graph as is in Mapper(UMAP) and Mapper(LargeVis). Similarly \textit{Bag} class is connected with \textit{T-shirt/Top} as compared to \textit{Ankle-Boot} class. The loop in ShapeVis visualization captures the similarity chain of the classes \textit{Dress}, \textit{T-shirt}, \textit{Pullover} and \textit{Coat}, which other approaches fail to capture. Another detail which ShapeVis captures more vividly is that images of sleeveless tees though labeled \textit{T-shirt} is clubbed with \textit{Dress} nodes because of its visual similarity to short dresses than the \textit{T-shirt} node. 

\paragraph{\textbf{Unlabeled datasets}}
For LiveJournal and GoogleNews vectors, no ground truth class label is available. Therefore, for comparing the visualization quality of ShapeVis to other approaches, we assign pseudo labels to each data point and then color each with its dominant pseudo label. For assigning a pseudo label, we run Louvain community detection \cite{louvain} on ShapeVis (Mapper) graph to partition it into segments, and each data point is assigned the label of the segment to which it belongs. Fig \ref{lj_and_w2v}, rows 1 and 2, show the visualization on LiveJournal dataset when pseudo labels are assigned using segments of Mapper(UMAP) and ShapeVis, respectively. We can see in Fig \ref{lj_and_w2v} row 1, that the segments found through ShapeVis correspond well with the segments in Mapper(UMAP) and Mapper(LargeVis). Similarly, segments of Mapper (UMAP) align with the segments of ShapeVis and Mapper(LargeVis). This shows that visualization obtained through our approach is qualitatively similar to Mapper with UMAP or LargeVis filter functions. We do not show Mapper(t-SNE) as we were unable to run t-SNE on these datasets because of its huge time-complexity. Fig \ref{lj_and_w2v}, rows 3 and 4, show the visualization on GoogleNews vectors. Both Mapper(UMAP) and Mapper(LargeVis) fail to bring any clear segmentation of dataset in the visualization. Moreover, we do not see any alignment between segments of different visualizations. We also compute the average cosine similarity between word-vectors belonging to the same segments for all three visualizations. For ShapeVis, Mapper(UMAP) and Mapper(LargeVis) average cosine similarity between words of a segment is 0.224, 0.186 and 0.132, respectively. Thus ShapeVis performs slightly better by this measure.

\begin{table}[t]
\centering
\scalebox{0.95}{
\setlength\tabcolsep{1.5pt}
\begin{tabular}{|c|c|c|c|c|}
\hline
\textbf{Dataset} & ShapeVis & \begin{tabular}{@{}c@{}}Mapper \\ (UMAP)\end{tabular} & \begin{tabular}{@{}c@{}}Mapper \\ (LargeVis)\end{tabular} & \begin{tabular}{@{}c@{}}Mapper \\ (t-SNE)\end{tabular} \\
 \hline
MNIST & 483.01 & 217.19 & 679.67 & 723.33 \\
FMNIST  & 340.09 & 200.97  & 603.43 & 543.69 \\
Word Vectors  & 1796.12 & 3116.14 &  5880.67 & NA \\
LiveJournal & 3351.24 & 3729.32 & 13804.14  & NA \\
\hline
\end{tabular}}
\caption{\footnotesize{Time comparison (in seconds) on different datasets of all approaches.}}
\label{table:time-compare}
\end{table}

\begin{figure}[t]
\centering
\scalebox{1.}{
    \includegraphics[width=2.5in,height=1.0in]{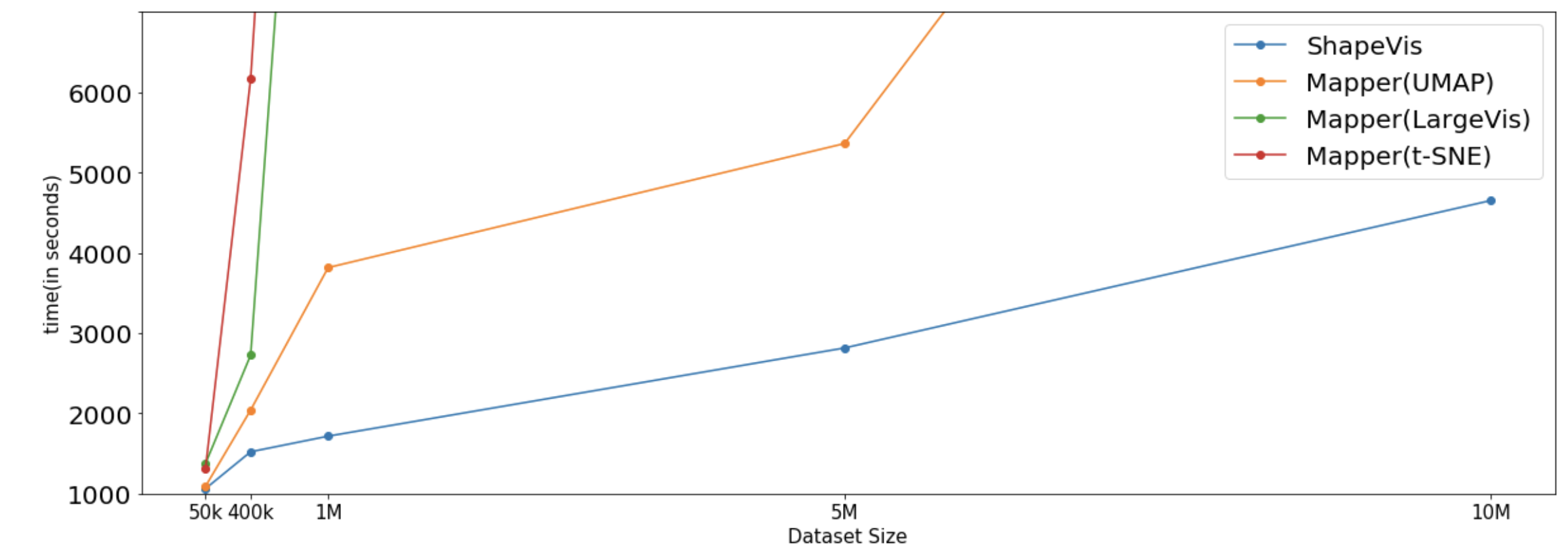} 
}
    \caption{\footnotesize{Running time (in seconds) of all approaches with increase in dataset size of points sampled from a uniform 25-dim sphere}}
    \label{fig:running-time}
\end{figure}

\subsection{Time Comparison}
We compare the running time of ShapeVis against other approaches on all the above mentioned datasets. All the results are executed on a machine with 48GB memory and 6 cores. For LiveJournal and GoogleNews vectors, we compare on 2 million and 1 million subsets respectively since UMAP returned memory overflow error on the complete dataset. Table \ref{table:time-compare} shows the running time of all approaches in seconds. We can see that ShapeVis significantly outperforms other methods for large datasets and is comparable on smaller datasets. 

We further analyze the scalability of ShapeVis with dataset size by running it on random samples of points from a uniform sphere of 25-dimension. Fig \ref{fig:running-time} shows the plot of running time (in seconds) vs the number of sampled points for all approaches. It shows that as the dataset size increases ShapeVis is more and more efficient as compared to Mapper with t-SNE, UMAP or LargeVis filter functions.


\section{Conclusion}
In this paper, we proposed ShapeVis, a graph-based visualization technique that aims to preserve the topological structure of data in the form of a summarized graph. The 2-step landmark sampling in ShapeVis helps it to scale to millions of data-points with consistency. Experiments on labeled real-world datasets show that ShapeVis captures the global relationship between different labels coherently. For unlabeled datasets, the visualization of ShapeVis is qualitatively similar to existing approaches. It captures the relationship between different local neighourhoods of data in a concise manner and scales with significantly lower running time. In the future, we aim to incorporate hierarchical visualization into ShapeVis. Although we show only high level details in the current visualization, it can be easily extended to interactively explore the segments of the graph at a finer scale.

\bibliographystyle{ACM-Reference-Format}
\bibliography{sample-base}




\end{document}